\documentclass[5p]{elsarticle}
\biboptions{authoryear}

\journal{Medical Image Analysis}
\usepackage{gensymb}
\usepackage{appendix}
\usepackage{graphicx}
\usepackage[utf8]{inputenc}
\usepackage[normalem]{ulem}
\usepackage{xspace} 
\usepackage{booktabs, multicol, multirow}
\usepackage[table]{xcolor}
\usepackage[T1]{fontenc}
\usepackage{relsize}
\usepackage{mathtools}
\usepackage{lipsum}

\usepackage{hyperref}
\usepackage{amsmath}
\usepackage{cleveref}
\usepackage{amsfonts}
\usepackage{siunitx}
\usepackage{algorithm}
\usepackage{array}
\usepackage{float}  
\usepackage{microtype}
\usepackage{subcaption}
\usepackage{xcolor}
\usepackage[switch]{lineno}
\usepackage{fmtcount} 
\usepackage{multirow}
\usepackage{textcomp}
\usepackage{algpseudocode}
\usepackage{chngcntr}
\newcolumntype{C}[1]{>{\centering\arraybackslash}m{#1}}

\begin{document}
\begin{frontmatter}
\title{EndoSLAM Dataset and An Unsupervised Monocular Visual Odometry and Depth Estimation Approach for Endoscopic Videos: Endo-SfMLearner}

\author[a]{Kutsev Bengisu Ozyoruk}
\ead{bengisu.ozyoruk@boun.edu.tr}

\author[a]{Guliz Irem Gokceler}

\author[a]{Gulfize Coskun}

\author[a]{Kagan Incetan}

\author[b]{Yasin Almalioglu}

\author[c,d,e]{Faisal Mahmood}

\author[f]{Eva Curto}

\author[f]{Luis Perdigoto}

\author[f]{Marina Oliveira}

\author[a]{Hasan Sahin}

\author[f]{Helder Araujo}

\author[i]{Henrique Alexandrino}

\author[g]{Nicholas J. Durr}

\author[h]{Hunter B. Gilbert}

\author[a]{Mehmet Turan\corref{cor1} }
\ead{mehmet.turan@boun.edu.tr}

\cortext[cor1]{Corresponding Author}
\fntext[fn2]{This work was supported by the Scientific and Technological Research Council of Turkey (TUBITAK) with grant 2232 - The International Fellowship for Outstanding Researchers.}

\address[a]{Institute of Biomedical Engineering, Bogazici University, Turkey}
\address[b]{Computer Science Department, University of Oxford, Oxford, UK }
\address[c]{Brigham and Women's Hospital, Harvard Medical School, Boston, MA, USA}
\address[d]{Cancer Data Science, Dana Farber Cancer Institute, Boston, MA, USA}
\address[e]{Cancer Program, Broad Institute of Harvard and MIT, Cambridge, MA, USA}
\address[f]{Institute for Systems and Robotics, University of Coimbra, Portugal}
\address[g]{Department of Biomedical Engineering, Johns Hopkins University, Baltimore, MD}

\address[h]{ Department of Mechanical and Industrial Engineering, Louisiana State University, Baton Rouge, LA, USA}
\address[i]{Faculty of Medicine, Clinical Academic Center of Coimbra, University of Coimbra, Coimbra, Portugal}


\begin{abstract} 
Deep learning techniques hold promise to develop dense topography reconstruction and pose estimation methods for endoscopic videos. However, currently available datasets do not support effective quantitative benchmarking. In this paper, we introduce a comprehensive endoscopic SLAM dataset consisting of 3D point cloud data for six porcine organs, capsule and standard endoscopy recordings as well as synthetically generated data. A Panda robotic arm, two commercially available capsule endoscopes, two conventional endoscopes with different camera properties, and two high precision 3D scanners were employed to collect data from eight ex-vivo porcine gastrointestinal (GI)-tract organs. In total, 35 sub-datasets are provided with 6D pose ground truth for the ex-vivo part: 18 sub-datasets for colon, 12 sub-datasets for stomach and 5 sub-datasets for small intestine, while four of these contain polyp-mimicking elevations carried out by an expert gastroenterologist. Synthetic capsule endoscopy frames from stomach, colon and small intestine with both depth and pose annotations are included to facilitate the study of simulation-to-real transfer learning algorithms. Additionally, we propound Endo-SfMLearner, an unsupervised monocular depth and pose estimation method that combines residual networks with spatial attention module in order to dictate the network to focus on distinguishable and highly textured tissue regions. The proposed approach makes use of a brightness-aware photometric loss to improve the robustness under fast frame-to-frame illumination changes that is commonly seen in endoscopic videos. To exemplify the use-case of the EndoSLAM dataset, the performance of Endo-SfMLearner is extensively compared with the state-of-the-art: SC-SfMLearner, SfMLearner and Monodepth2. The codes and the link for the dataset are publicly available at {\color{blue}\url{https://github.com/CapsuleEndoscope/EndoSLAM}}
. A video demonstrating the experimental setup and procedure is accessible through   {\color{blue}\url{https://www.youtube.com/watch?v=G_LCe0aWWdQ}}.
\end{abstract}

\end{frontmatter}

\section{Introduction}
\label{sec:intro}
Gastrointestinal(GI) cancers affect over 28 million patients annually, representing about 26\% of the global cancer incidence and 35\% of all cancer-related deaths \citep{globalcancerfacts}. Besides, GI cancer is the second deadliest cancer type with reported 3.4 million GI related deaths  globally in 2018 \citep{Arnold2020Global}. Direct visual inspection(DVI) of these cancers is the simplest and most effective technique for  screening. Esophagogastroduodenoscopy (EGD) and colonoscopy are used to visualize gastrointestinal diseases specifically in colon and rectum while capsule endoscopy (CE) is preferred for small bowel exploration \citep{2014WCE}. 

An endoscopic gastro-intestinal procedure analysis hold by iData Research reveals that over 19 million colonoscopies are performed annually, as reported in 2017, a tremendous contribution to the 75 million endoscopies applied each year in the United States \citep{iDataResearch}.
Specifically, the malignant tumors developed in the small intestine like Adenocarcinoma, Intestinal Lymphoma, Leiomyosarcoma, and metastatic malignancy from lung or breast are severe diseases, mostly resulting in death. Among these, the small bowel involving polyposis syndromes include Familial Adenomatous Polyposis, generalized Juvenile polyposis, Peutz-Jeghers and Cronkhite-Canada syndromes are the most mortal types. The diagnosis of these polyps and small-bowel tumors are challenging due to rarity of lesions, lack of common symptoms across patients, and variety of the symptoms \citep{smallBowel2009}. In these cases, differential diagnosis from blood tests and symptoms alone are not sufficient, and visual examination through capsule endoscopy can provide valuable information. After visual confirmation of any feature of diagnostic importance, ``where is it?'' arises as natural question. In the following subsection, we overview the related work from literature which are all motivated by this critical question.  

\subsection{Related work}
\label{ssec:relatedWork}
The direction of arrival estimation based localization techniques such as radio frequency based signal triangulation \citep{dey2017wireless}, received signal strength \citep{ shah2006development}, electromagnetic tracking \citep{son20155}, x-ray \citep{kuth2007method} and positron emission markers \citep{than2014effective} have been widely investigated in robotics.
In capsule endoscopes, visual information has been provided which drives the attention to the development of vision-based odometry and simultaneous localization and mapping (SLAM) systems, either to remove the need for added hardware for pose sensing or to provide additional information for 3D tracking. While current capsules are propelled by the peristaltic motion of the GI tract, active capsule endoscopes hold promise to provide drug delivery and biopsy \citep{ciuti2016frontiers}. Vision-based SLAM is of utmost importance to enable these functions and other forms of complementary situational awareness in decision support and augmented reality systems \citep{simaan2015intelligent}.
With the rise of deep learning techniques \citep{endoVOTURAN2018}, public datasets enabling a broader research community to work on the localization and mapping problems became crucial \citep{Kvasir2017,MICCAI2017,HyperKvasir} in medical image analysis. Several datasets are available to support research and development of a variety of advanced diagnostic features across a wide range of tasks, including segmentation, disease classification, tissue deformation and motion detection, and depth estimation. Some of them are available in the context of endoscopy which are overviewed in Table~\ref{Tab:dso} and explained more in detail in Appendix B. 
\begin{figure}[thb] 
  \centering
  \includegraphics[width = \columnwidth]{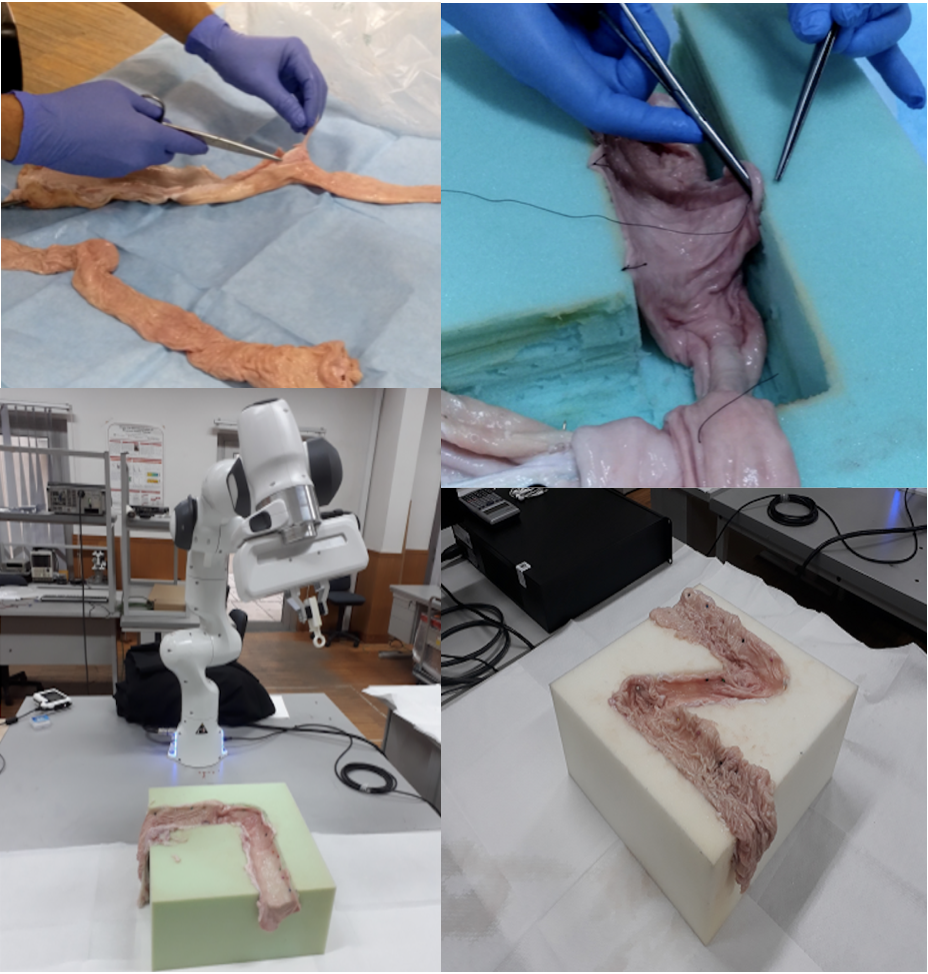}
  \caption{
  \textbf{Experimental Setup.} The pre-harvested and cleaned porcine colons are cut and sewn to L and Z-shaped scaffolds by a health practitioner.  A data belt, copper wiring and wireless transmitter are utilized for the data transfer between the receiver and wireless capsule endoscope(WCE). A specially designed capsule holder is attached to the robotic arm for high precision pose ground truth (see Fig. \ref{fig:equipment}). All experiments for the ex-vivo part of the dataset are conducted in dark room to mimic the real capsule endoscopy procedure.}

\label{fig:expdemon}
\end{figure}

{\centering
\begin{table*}[!h]
\caption{\textbf{Dataset Survey.} An overview of existing datasets for disease classification, polyp recognition, segmentation, pose tracking and depth estimation. The size of each dataset in terms of number of images, and  corresponding organs are also listed. The datasets, collected via capsule endoscopy, standard endoscopy and laparoscopy, are denoted by $^{\diamond}$, $^{\dagger}$ and $^{\star}$, respectively. } \resizebox{\textwidth}{!}{
{\small\centering
\begin{tabular}{p{4cm} p{4cm}lll}

     {Dataset Name} & {Findings} & {Organs} & {Tasks} & {Size} \\
     \midrule\midrule[.1em]

     {Kvasir-SEG$^{\dagger}$} & Polyps & Colon & Segmentation & 1,000 \\
     \midrule[.1em]
     \multirow{2}{4cm}{Kvasir$^{\dagger}$ \citep{Kvasir2017}}
      &{Z-line, pylorus, cecum, esophagitis, polyps,  ulcerative colitis, dyed}
      & Colon 
      &{Disease detection}
      &{6,000} \\ 
      \cmidrule(lr){2-5}
      &{Lifted polyps and dyed resection margins}
      & Colon
      &{Segmentation}
      &{2,000}\\
    \midrule[.1em]    
    \multirow{3}{4cm}{Hamlyn  Centre Datasets$^{\dagger \star}$}
      &{Polyp}
      &{Colon}
      &{Segmentation}
      &{7,894}\\
    \cmidrule(lr){2-5}
      &{-}
      &{Kidney}
      &{Disparity}
      &{40,000}\\
    \cmidrule(lr){2-5}
      &{Polyp}
      &{Colon}
      &\vtop{\hbox{\strut Polyp recognition}\hbox{\strut Localisation}}
      &{2,000}\\
    \cmidrule(lr){2-5}
      &{-}
      &{Liver, ureter, kidney, abdomen}
      &\vtop{\hbox{\strut Tissue deformation }\hbox{\strut Tracking}}
      &{-}\\
\midrule[.1em]      
     KID  Dataset$^{\diamond}$ \citep{pmid28580415} & Angioectasias, apthae, chylous cysts and polypoid, vascular and inflammatory lesions & Small Bowel and colon & Classification & 2,448\\
\midrule[.1em]
 NBI-InFrames$^{\dagger}$ \citep{NBIInFrameDataset}& Angioectasias, apthae, chylous cysts and polypoid & Larynx & Classification & 720\\
\midrule[.1em]
    { EndoAbs$^{\dagger}$ \citep{endoabsdataset}} & -  & Liver, kidney, spleen & Classification & 120\\
\midrule[.1em]
    { ASU-MAYO Clinic$^{\dagger}$} & Polyp & Colons & Segmentation & 22,701\\        

\midrule[.1em]
    { ROBUST-MIS Challenge$^{\star}$} & Rectal cancer & Abdomen & Segmentation & 10,040\\ 

\bottomrule

\end{tabular}
}
\label{Tab:dso}
}

\end{table*}
}

\subsubsection{Survey for Depth and Pose Estimation}
\label{sec:depthPoseEstimation}

Depth estimation from a camera scene and visual odometry are very challenging and active problems in computer vision. Various traditional multi-view stereo \citep{Hartley:2003:MVG:861369} methods such as structure from motion \citep{Wu11multicorebundle,Leonard2018} and SLAM \citep{grasa2013visual} can be used to reconstruct 3D map based on the feature correspondence. However, their performances are still far from being perfect especially for endoscopic images suffering from lack of distinguishable feature. Despite the recent advances in image processing, colonoscopy remains as complicated procedure for depth estimation because of monocular camera with insufficient light source, limited working area and frequently changing environment due to the contractions of muscles. In that regard, deep-learning based methods have been applied for monocular depth estimation \citep{Liu2020,NIPS2014_5539,Liu2016}. CNN-based depth estimation methods have shown promising performance on a single image depth inference despite the scale inconsistency \citep{DBLPLainaRBTN16}. Nevertheless, using CNN in a fully supervised manner is challenging for endoscopy since dense depth map ground truth that correspond directly to the real endoscopic images are hard to obtain. Even if the labeled dataset is provided, patient-specific texture, shape and color make difficult to get generalizable results without a large amount of ground truth. These issues are mostly overcome by either synthetically generated data or the simultaneous depth and pose estimation methods where the output of pose network supervises the depth network instead of human expert annotations \citep{turan2018unsupervised, Lu2019}. Mahmood et al. propose unsupervised reverse domain adaptation framework to avoid these annotation requirements which is accomplished by adversarial training removing patient specific details from real endoscopic images while protecting diagnostic details \citep{2018mahmood}.  In \citep{DBLP_CRF2018}, the monocular depth estimation is formulated as conditional random fields learning problem and CNN-CRF framework that consists of unary and pairwise parts are introduced as domain adaptable approach. Several self-supervised methods related with the single-frame depth
estimation have been propounded in the generic field of computer vision \citep{DBLPGargBR16,Zhang2020,Yin2018GeoNetUL}. However, they are not generally applicable
to endoscopy because of inter-frame photometric constancy assumptions of these works which is broken by the frequently appearing inconsistent illumination profile in endoscopic videos. The jointly moving camera and light source cause  the appearance of the same anatomy differ substantially with varying camera poses, especially for tissue regions close to the camera surface. This might give rise to the network to get stuck in a local minima during training, specifically for textureless regions where extracting reliable information from photometric appearance is extremely difficult \citep{Chen2019}. There are also studies solely focusing on monocular localization problems utilized by CNN \citep{DBLPMagicVO2018,EasyChair_413}. Unlike traditional artificial neural networks, Turan et al. use RCNN which is able to process arbitrarily long sequences by its directed cycles between the hidden units and infer the correlative information across frames  \citep{endoVOTURAN2018}. However, estimating a global scale from monocular images is inherently ambiguous \citep{NIPS2014_5539}. Despite all efforts, visual odometry is
insufficient in real-time localization and vSLAM methods come on the scene as a solution which can be tested only via a comprehensive vSLAM dataset with accurate ground truths. In the work of Mountney et al., a vSLAM method based on Extended Kalman Filter SLAM (EKF-SLAM) is used for localization and soft tissue mapping where sequential frames acquired by moving stereo endoscopes \citep{mountney2006simultaneous}. In robotic surgical systems such as da Vinci\texttrademark, real-time 3D reconstruction methods have been applied and validated on phantom models \citep{stoyanov2010real,lin2013simultaneous}. Lin et al. adopt and extend Parallel Tracking and Mapping (PTAM) method to detect deformations on a non-rigid phantom to create 3D reconstruction of intestine model and to track endoscope position and orientation \citep{lin2013simultaneous}. Some other works are focused on more commonly used monocular endoscopes. Mirota et al. generate a 3D reconstruction from endoscopic video during sinus surgeries by using feature detection and registered data from CT scan tracking endoscope location \citep{mirota2011system}. In \cite{grasa2013visual}, another monocular vSLAM method is used to provide real-time 3D map of the abdominal cavity for hernia repair interventions. Apart from standard endoscopes, vSLAM techniques have also been used in capsule endoscopy \citep{Chen2019SLAMEE,turan2017non}. 
A robust and reliable SLAM module is indispensable for next-generation capsule robots equipped with the functionalities including biopsy, drug delivery and automated polyp detection \citep{Turan3Drecons2017}, but several technical challenges such as low frame rate and low resolution due to space limitations make this need tough to meet. Specular reflections from extracellular fluids and rapidly changing environment due to peristaltic motions are further examples of inherent challenges. Those problems have motivated the exploration of deep learning based approaches that eschew complex physical models which ends up with the necessity of huge amount of dataset. 
\begin{figure*}[!ht]
    \centering
    \includegraphics[width = 2 \columnwidth]{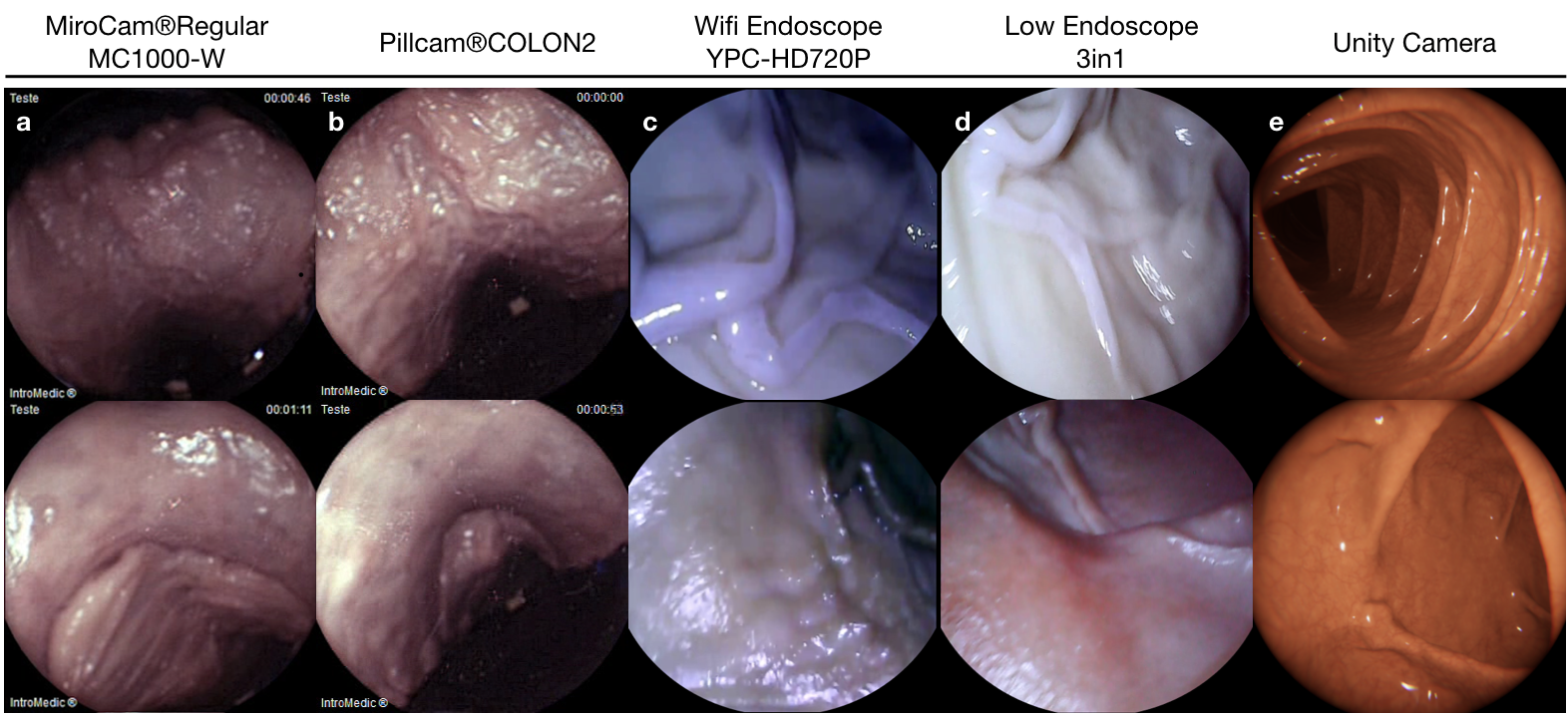}
    \caption{\textbf{Sample frames from EndoSLAM Dataset.} Images are acquired by \textbf{a} MiroCam capsule endoscope, \textbf{b} Frontal camera of a PillCam, \textbf{c} HighCam, \textbf{d} LowCam, and \textbf{e} virtually generated UnityCam. The ex-vivo part of the dataset offers opportunity to test the robustness of pose estimation algorithms with images coming from various endoscope cameras. Since EndoSLAM dataset contains real and simulated frames, it is also a suitable platform to develop domain adaptation algorithms.    
    }
    \label{fig:samptraj}
\end{figure*}

\subsection{Contributions}
\label{ssec:contributions}
In this work, we introduce the EndoSLAM dataset, a dedicated dataset designed for the development of 6-DoF pose estimation and dense 3D map reconstruction methods. The dataset is recorded using multiple endoscope cameras and ex-vivo porcine GI organs belonging to different animals and is designed to meet the following major requirements for scientific research and development of endoscopic SLAM methods:
\begin{itemize}
\item  Time-synchronized, ground-truth 6 DoF pose data
\item  High precision, ground-truth 3D reconstructions
\item  Multiple organs from multiple individuals
\item  Images from cameras with varying intrinsic properties
\item  Image sequences with differing native frame rates
\item Images acquired from different camera view angle such as perpendicular, vertical and tubular
\item  Images under a variety of lighting conditions
\item  Distinguishable features of diagnostic significance (e.g. presence/absence of polyps).
\end{itemize}
In addition to the experimentally collected data, synthetically generated data from a 3D simulation environment is included to facilitate study of the simulation to real-world problems such as domain adaptation and transfer learning.
One of the biggest disadvantages of deep learning techniques is the fact that large networks need massive amounts of domain-specific data for
training. Research in recent years has shown that large amounts of synthetic data can improve the performance of learning-based vision algorithms and can ameliorate the difficulty and expense of obtaining real data in a variety of contexts. However, due to the large gap between simulation data and real data, this path needs domain adaptation algorithms to be employed. With the synthetically generated data from Unity 3D environment, we aim to provide a test-bed to overcome the gap between simulation and real endoscopic data domain. 

In addition to the EndoSLAM dataset, we propose an unsupervised  depth and pose estimation approach for endoscopic videos based on spatial attention and brightness-aware hybrid loss. The main idea and details of the proposed architecture are depicted in Fig. \ref{fig:EndoSfM}. Our main contributions are as follows:
\begin{itemize}
\item Spatial Attention-based Visual Odometry and Depth-Estimation: We propose a spatial attention based ResNet architecture for pose estimation optimized for endoscopic images.
\item Hybrid Loss: We propose a hybrid-loss function which is specifically designed to cope with depth of field related defocus issues and fast frame-to-frame illumination changes in endoscopic images. It collaboratively combines the power of brightness-aware photometric loss, geometry consistency loss, and smoothness loss.
\end{itemize}
Rest of the paper is organized as follows: 
Section \ref{sec:data} describes the experimental setup and gives details about the specifications of the devices used for dataset recording, introduces the overall datatree structure. In Section \ref{sec:EndoSfM}, Endo-SfMLearner is described.  In Section \ref{sec:analys}, various use-cases of the EndoSLAM dataset are exemplified by benchmarking the Endo-SfMLearner and the state-of-the-art monocular depth and pose estimation methods SC-SfMLearner \citep{bian2019unsupervised}, SfMLearner \citep{8100183} and Monodepth2 \citep{2018monodepth2}. Besides, fully dense 3D map reconstruction is exemplified using EndoSLAM dataset and Endo-SfMLearner. Finally, Section \ref{sec:discussion} discusses the future plans and offers some concluding remarks.

\section{Dataset Shooting}
\label{sec:data}
In this section, we will introduce experimental setup, procedure and detailed structure of EndoSLAM dataset.
\subsection{Experimental Setup} 
\label{sec:setup}
The experimental setup was specifically designed to support the collection of endoscopic videos, accurate 6-DoF ground truth pose, organ shape and topography data. 
The essential components are four endoscope video cameras (see Fig. \ref{fig:equipment}i, j, l, m), a robotic arm to track the trajectory and quantify the pose values (see Fig. \ref{fig:equipment}a), and high precision 3D scanners for ground truth organ shape measurement (see Fig. \ref{fig:equipment}g-h). All of the equipment are illustrated in Fig. \ref{fig:equipment}.
As per camera devices, MiroCam\textsuperscript{\textregistered} (see Fig. \ref{fig:equipment}m) and Pillcam\textsuperscript{\textregistered} COLON2 (see Fig. \ref{fig:equipment}l) capsule endoscope cameras and two other cameras (HighCam and LowCam) representative of conventional endoscope cameras (see Fig. \ref{fig:equipment}i-j) were employed. Their specifications are as follows:  
\begin{itemize}
    \item MiroCam\textsuperscript{\textregistered} Regular MC1000-W endoscopic video capsule: 320$\times$320 image resolution, 3 fps frame rate, 170$\degree$ field of view, 7 - 20 mm depth of field, 6 white LED's (Fig. \ref{fig:equipment} m). \citep{MirocamVsPilcam2012}
    \item Pillcam\textsuperscript{\textregistered} COLON2 double endoscope camera capsule: 256$\times$256 each camera, 4 fps to 35 fps variable frame frate, 344$\degree$ field of view (172$\degree$ each camera), 4 LEDs (each camera), Fig. \ref{fig:equipment}l \citep{MirocamVsPilcam2012}.
    \item High Resolution Endoscope Camera (YPC-HD720P): 1280$\times$720 image resolution, 20 fps frame rate, 120$\degree$ field of view, 4-6 cm depth of field, 6 adjustable white LEDs, Fig. \ref{fig:equipment}i.   
    \item Low Resolution Endoscope 3 in 1 Camera: 640$\times$480 image resolution, 20 fps frame rate,  
    130$\degree$ field of view, 3-8 cm depth of field, 6 adjustable  LEDs, Fig. \ref{fig:equipment}j. 
\end{itemize}

\begin{figure*}[t]
\centering
\includegraphics[width=2\columnwidth]{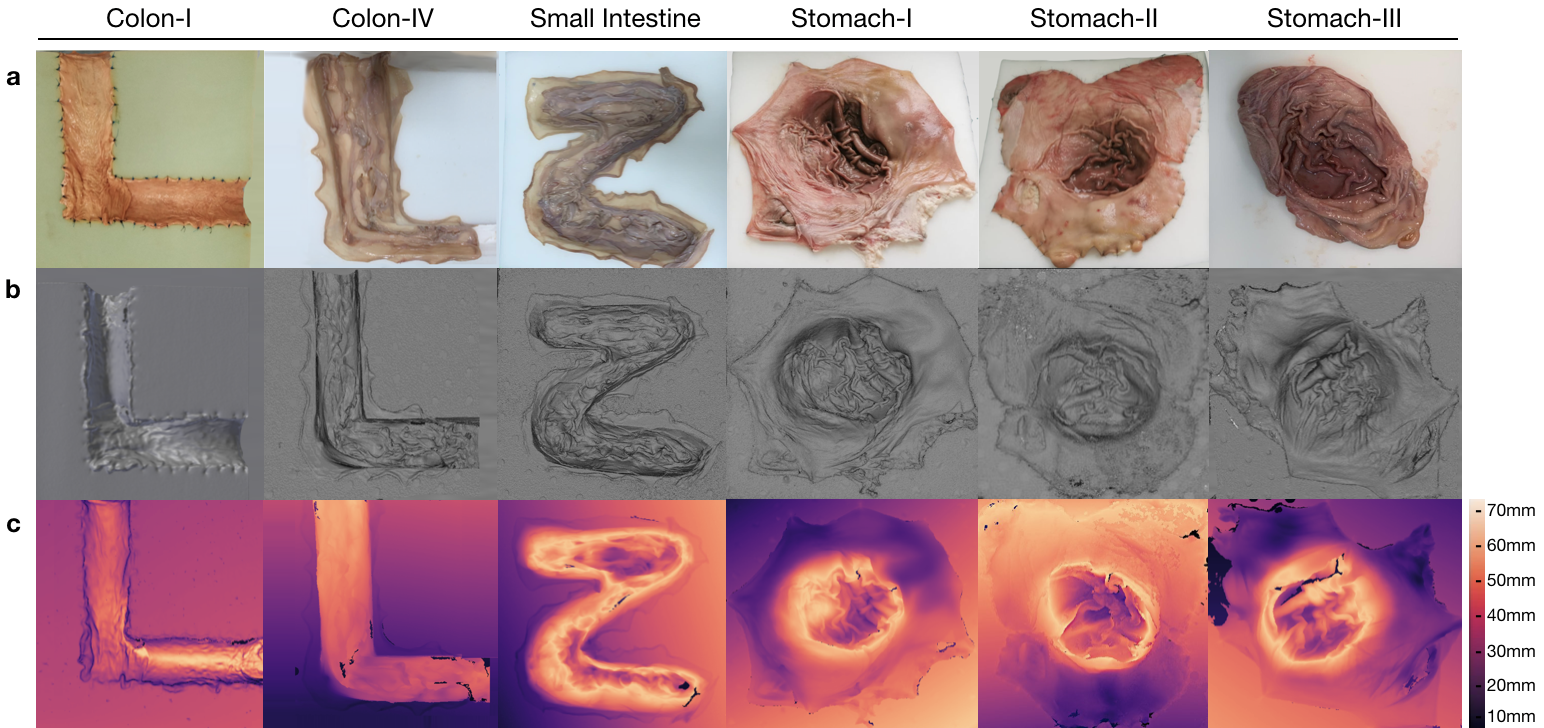}
\caption{\textbf{3D-Scanner images for EndoSLAM Dataset.} 
3D-scanner images for six organs which are fixed to scaffolds that were cut in O, Z and L shapes to mimic the GI-tract path through the ascending  colon to  the  transverse  colon. \textbf{a} RGB images of scanned organs.  
\textbf{b} Corresponding 3D reconstruction from .ply files for organs recorded via  3D Scanner. \textbf{c} Heatmap reconstructions for depth values by means of the Computer Vision Toolbox of Matlab. 3D point cloud data from two colons, one small intestine and three stomachs from different individuals make dataset appropriate not only development for 3D reconstruction algorithms but also for transfer learning problems.}
\label{fig:gt}
\end{figure*}

Ground truth geometries of the organs were acquired via two commercially-available 3D scanners, the Artec 3D Eva and Shining 3D Einscan Pro 2x (see Fig. \ref{fig:equipment}g-h). 3D models of organs were reconstructed as in Fig. \ref{fig:gt}   and the  depth distribution histograms for corresponding organs are given in Fig. \ref{fig:depthhisto}. 
Relevant performance specifications of the 3D scanners are as follows:
\begin{itemize}
\item Artec 3D Eva: $\pm$0.5 mm 3D resolution, $\pm$0.1 mm 3D point accuracy, $\pm$0.03\%  3D accuracy over 100 cm distance \citep{Eva3D}. 
\item Shining 3D EinScan Pro 2x: 0.2-2mm point distance; 
    $\pm$0.5 mm 3D resolution, $\pm$0.05 mm 3D point accuracy, $\pm$0.03\%  3D accuracy over 100 cm distance \citep{Einscan3D}. 
\end{itemize}

In the rest of the paper, we will call MiroCam\textsuperscript{\textregistered} Regular MC1000-W endoscope video capsule as MiroCam, Pillcam\textsuperscript{\textregistered} COLON2 as PillCam, High Resolution Endoscope Camera (YPC-HD720P) as HighCam and Low Resolution Endoscope 3 in 1 Camera as LowCam for brevity. 
Franka Emika Panda robotic arm with 7 degree of freedom and 0.1mm pose repeatability precision is utilized to provide trajectory stabilization and ground-truth pose recording of the camera motions. The robot controls the motion of the cameras in hand-guided mode and supplies pose recording at high-frequency, 1kHz (see Fig. \ref{fig:equipment} a).

\subsection{Dataset Collection}
The ex-vivo part of dataset consists of a total of 42,700 frames. Of these, 21,428 images are from the HighCam, 17,978 images are from the LowCam, 239 images are from the PillCam and 3,055 images are from the MiroCam.
The reader is referred to Fig.~\ref{fig:samptraj} and Fig.~\ref{fig:samptraj2} for the illustration of recorded frames. 
Fig.~\ref{fig:folder} shows the overall data tree structure. \texttt{Frames} folder contains recorded endoscopic images in a given trajectory and Table~\ref{table:coltraj} summarizes the trajectory classes based on tumor comprising for each organ. Trajectories for each organ can be found in the folder \texttt{Poses} in (.csv) and (.txt ) format with 6D pose coordinates: four orientation parameters (x; y; z; w) in quaternions and three absolute position parameters (x; y; z) in meters. \texttt{Calibration} folder contains intrinsic-extrinsic calibration data of cameras in .mat extensions and calibration sessions. \texttt{3D\textunderscore Scanners} folder consists of reconstructed 3D figures and point cloud data for six organs with a size of 23.2 GB in total, the reader is referred to the Table \ref{tab:3Dpoint} for the detailed point cloud distributions.

\subsection{Experimental Procedure}

'L'-shaped, 'Z'-shaped semi-cylindrical and 'O' shaped semi-spherical scaffolds were cut in rectangular high-density solid foams with dimensions $ 30 \times 30 \times 17 $ cm to be used as substrate for colons, small intestine and stomachs, respectively. The shape for colons mimics the GI-tract path through the ascending colon to the transverse colon. Cleaned porcine organs were cut and sewn to the foam by a practitioner, see Fig. \ref{fig:expdemon}. 

All the capsules require a specific recorder to be worn by the patient. PillCam and MiroCam differ in multiple features and, in particular, PillCam transmits video wirelessly whereas MiroCam transmits video via a wired connection. For both PillCam and MiroCam, the capsules are placed into a 3D-printed, non-conducting holder attached to the robotic arm.  HighCam and LowCam were also used in place of the capsule cameras to record endoscopic images in a similar set-up but with different organs, Colon-IV, Small Intestine, Stomach-I, Stomach-II, and Stomach-III, which were sewn on white scaffold. 

Recordings were made in a dark room with green and white background solid foams to create luminance and color contrast between the GI tract and environment. The orientations of the cameras (capsule and conventional) throughout the procedure are mostly along the longitudinal axis of the semi-cylindrical and semi-spherical surfaces. In most of the trajectories the capsule endoscopes do not contact with the tissue, whereas all cameras never contact with the tissue for all trajectories. In all the experiments the robot end-effector was driven by hand, with speeds ranging mostly between 16.76 mm/s to 25.97 mm/s with a peak speed of 286.68 mm/s and with accelerations mostly ranging from 279.254 mm/$s^2$  to 519.361 mm/$s^2$ with a peak acceleration of 14,680.15 mm/$s^2$. Alternating speed and accelerations are quite important in terms of SLAM evaluations, since performance of SLAM methods in general are significantly dependent on the complexity of the trajectories. In that regard, we performed detailed quantitative analysis of robot motion, see Table \ref{tab:trajSpeedAccHigh}, \ref{tab:trajSpeedAccLow}.

Experimental equipment calibration can be grouped into three parts as: camera calibration, the hand-eye transformation from robot to camera, and temporal synchronization between the camera frames and robot pose measurements. Each camera was calibrated against a pinhole camera model with non-linear radial lens distortion by Camera Calibration Toolbox MATLAB R2020a based on the theory of Zhang~\citep{zhang1999flexible} with the chessboard images illustrated in Fig. \ref{fig:MirocamCalib2mm}. The hand-eye transformation between the robot end-effector coordinate frame and the camera frame was estimated with the procedure of Tsai and Lenz \citep{HandEyeCalibration1989} and the resultant transformation matrices are given in Table \ref{tab:hand_eye}. Finally, temporal synchronization was performed by correlating camera motion computed by optical flow with velocity measurements from the robot. Table \ref{tab:TempSyncHighLow} and Table \ref{tab:TempSync} shows the correspondence between the start frame of each sequence and the matching sampling instant of the robot pose data. Further details of the calibration procedures are given in Appendix C and D.

\subsection{Dataset Augmentation}
\label{ssec:datasetaugmentation}
For the purpose of studying the robustness of SLAM algorithms against artefacts, the functions that are changing the property and quality of images were designed and exemplified in Fig. \ref{fig:lowCam_effects}. The transformations include resizing, Gaussian blur, fish-eye distortion, depth-of-field simulation via shift-variant defocus blurring, and frame-rate variation. The resize, vignetting and gaussian blur transformations were implemented with the opencv-python library (version 4.2.0.32), fish-eye distortion with the Pygame library (version 1.9.6), and depth-of-field with Matlab (version R2020a) \citep{depthOffield2020}. All codes for dataset augmentation functions are available in {\color{blue}\url{https://github.com/CapsuleEndoscope/EndoSLAM}}.

\subsection{Synthetic Data Generation}
In addition to real ex-vivo part of EndoSLAM dataset, we have generated synthetic capsule endoscopy frames to facilitate the study of simulation-to-real transfer of learning-based algorithms. The simulation environment, VRCaps \citep{ncetan2020vrcaps}, provides synthetic data which is visually as well as morphologically realistic. The platform was built with the use of real computed tomography (CT) images in DICOM format for topography and endoscopic images in RGB format for texture assignment. A cinematic rendering tool mimicking the effects in real capsule endoscopy records such as specular reflection, distortion, chromatic aberration, and field of view was used in order to obtain more photo-realistic images. Operating the virtual capsule inside the virtual 3D GI tract, we have recorded three sample endoscopic videos that containing 21,887 frames from colon, 12,558 frames from small intestine and 1,548 frames from stomach with pixel size of 320x320 and having both positional and pixel-wise depth ground truth.

 \section{Endo-SfMLearner} 
\label{sec:EndoSfM}
Recent works have proven that CNN-based depth and ego-motion estimators can achieve high performance using unlabelled monocular videos. However; static scene assumption, scale ambiguity between consecutive frames, brightness variety which basically stems from shallow depth-of-field and the organ tissues exhibiting non-lambertian surface property which are non-diffusely reflecting light particles make difficult to provide both locally and globally consistent trajectory estimations. We are proposing Endo-SfMLearner framework which specifically  addresses these gaps.

Endo-SfMLearner jointly trains a camera pose and depth estimation networks from unlabeled endoscopic dataset. Our method proposes two solutions to the light source rooted problems in depth and pose estimation. First proposed solution is to equate brightness conditions throughout the training and validation sets with brightness transformation function and the other is to weight the photometric loss with the brightness coefficient to punish the depth estimation with higher cost under different enlightenment conditions. Apart from these, we are using geometry consistency loss for scale-inconsistency between consecutive frames caused by alternating distances between camera and organ tissue. In principal, we convert the predicted depth map in one frame to 3D space, then project it to the consecutive frame using the estimated ego-motion, and minimize the inconsistency of the estimated and the projected depth maps. This implicitly compels the depth network to predict geometrically consistent (i.e. scale-consistent) results over consecutive frames. With iterative sampling and training, the frame-to-frame consistency can eventually propagate through the entire video sequence. As the
scale of depths is strictly linked to the scale of ego-motions, the ego-motion network can
estimate scale-consistent relative camera poses over consecutive pairs. The detailed network architecture for both depth and pose networks will be introduced in the following subsections.
\begin{figure*}[!t]
\centering
\includegraphics[width=1.97\columnwidth]{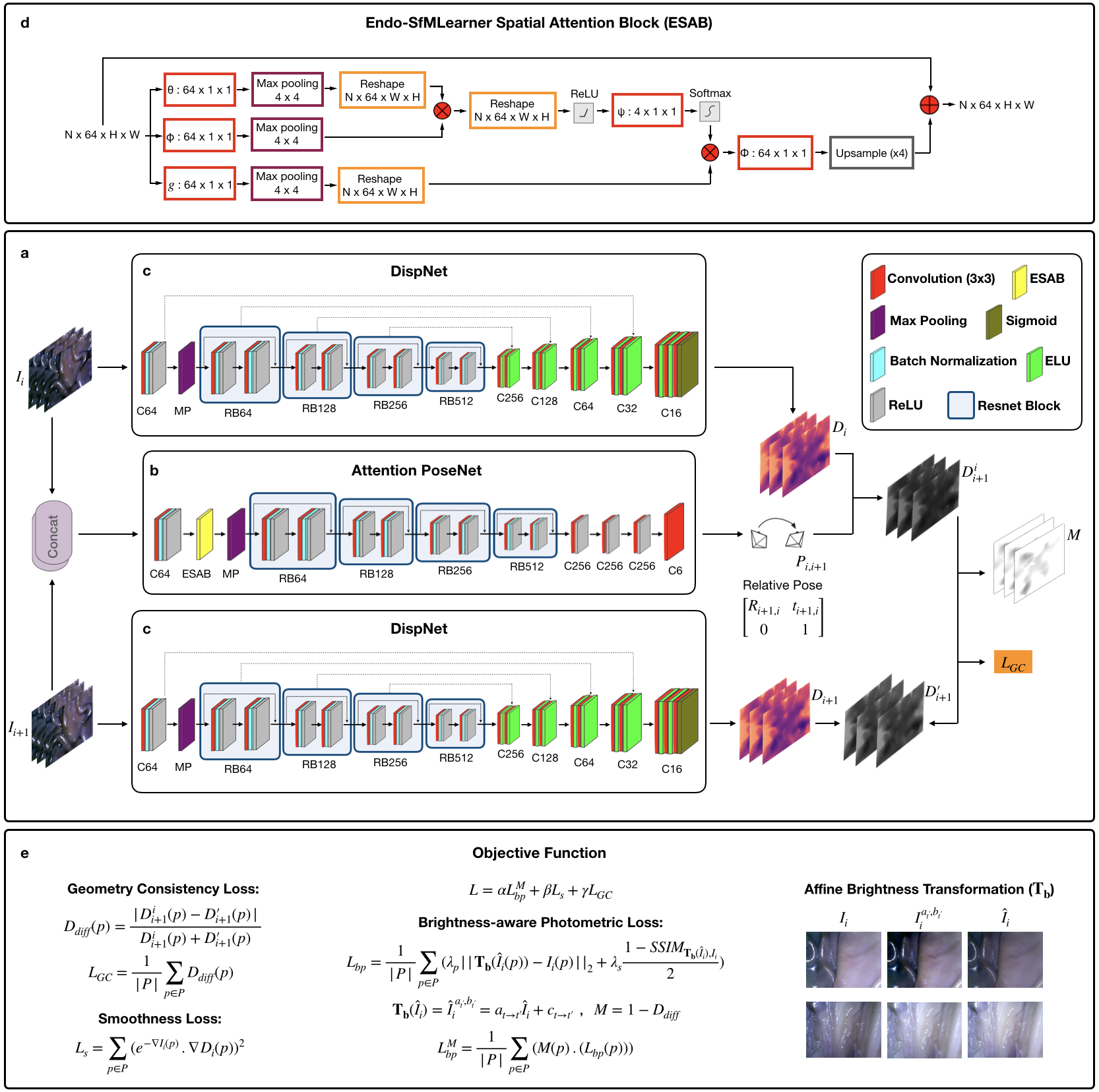}
    \caption{\textbf{Endo-SfMLearner Architecture Overview}. \textbf{a} Overall system architecture of Endo-SfMLearner framework. First of all, two consecutive unlabeled images ($I_{i},I_{i+1}$) are  fed into depth network separately and their corresponding dense disparity maps are predicted ($D_i,D_{i+1}$). PoseNet outputs the relative 6D camera poses $P_{i,i+1}$ for the same snippet. Reference images, $\hat I_i$, are synthesized with predicted depth and pose by warping the source image $I_{i+1}$. The difference between $\mathbf{T}_b(\hat I_i)$ and $I_i$ master the brightness-aware photometric loss. To deal with violation of geometric assumptions in image reconstruction (due to insufficiency of endoscope cameras), we also use geometry consistency loss which takes into account difference between warped $D_{i+1}^{i}$ and interpolated $D_{i+1}'$ pixel-wise disparity estimation. \textbf{b} Attention-PoseNet open form. The encoder part of the network consists of four basic ResNet blocks with spatial attention module in between ReLU and maxpooling layer. The decoder consists of four convolution layers each followed by ReLU activation function except last layer. \textbf{c} DispNet open form. DispNet encoder share similar structure with PoseNet encoder except ESAB block and skip connections. Decoder consists of five layers each consists of two convolution layers followed by ELU activation function. With the final convolution operation followed by sigmoid activation function, it outputs the dense disparity map from single image. \textbf{d} The flow diagram of Endo-SfMLearner spatial attention block operations. The non-local operator deduces the relative weights of all positions on the feature maps which measures the input covariance as a degree of tendency between two feature maps at different channels. For GPU memory usage efficiency which is crucial in global attention applications, max-pooling operations are integrated to the block operations. Thanks to the attention mechanism, PoseNet selectively focuses on texture details for more accurate pose and orientation estimation. \textbf{e} Endo-SfMLearner objective function. We are using a weighted sum of brightness-aware photometric loss, smoothness loss and geometry consistency loss as an overall learning objective. Affine brightness transformation function is utilized to equate the illumination conditions in between reference and target image before calculating SSIM and their pixelwise channel differences.  } 
    
    \label{fig:EndoSfM}
\end{figure*}
\subsubsection{Endo-SfMLearner Depth Network (DispNet)}
\label{ssec:DispNet}
The depth network which consists of encoder and decoder parts takes single image $I_i$ as input and gives output the corresponding disparity map $D_{i}$. For the sake of brevity, hereinafter we refer the batch normalization layer as BN, Rectified Linear Unit activation function as ReLU, exponential linear unit as ELU. Let RBk denote basic ResNet Block with \textit{k} filters and \texttt{Ck} is 3x3 convolution layer with \textit{k} filters. \texttt{C$_e$k}, \texttt{C$_s$k} and \texttt{C$_r$k} stand for \texttt{Ck} followed by ELU, sigmoid and ReLU, respectively.
\begin{itemize}
    \item \textbf{DispNet Encoder}
     DispNet encoder initializes with C64 with 7 kernel size, 2 stride and 3 padding followed by BN, ReLU activation function with a slope of 0.01 and max pooling operation with kernel size 3 and stride 2. Then, four ResNet basic blocks: RB64, RB128, RB256 and RB512 finalize the encoder structure. 
     Each ResNet basic block consists of Ck(3x3), BN, ReLU, Ck, BN and ReLU with skip connection. 
    \item \textbf{DispNet Decoder}
    DispNet decoder consists of five layers each consists of two convolution operations as follows:
    
    \texttt{C$_e$256}(x2)  -\texttt{C$_e$128}(x2)- \texttt{C$_e$64}(x2) - \texttt{C$_e$32}(x2) -  \texttt{C$_e$16}(x2)-\texttt{C$_s$16} 
\end{itemize}

To establish the information flow in between encoder and decoder, we are building skip connections from $i^{th}$ to $n-i^{th}$ layer where n indicating the total number of layers and i$\in$\{0,1,2,3\}, the reader is referred to Fig. \ref{fig:EndoSfM} c to overview.
\subsubsection{Endo-SfMLearner Pose Network (Attention PoseNet)}
\label{ssec:PoseNet}
The pose network takes the consecutive image tuples $(I_i,I_{i+1})$ as input by superposing and outputs the relative 6-DoF pose, $P_{i,i+1}$. 
\begin{itemize}
    \item \textbf{Attention PoseNet Encoder}
    We have integrated attention module to the encoder of PoseNet between ReLU and maxpooling layers.
    
    \texttt{C64}-BN-ReLU-\textbf{ESAB}-RB64-RB128-RB256-RB512

    \item \textbf{Attention PoseNet Decoder}
    
    \texttt{C$_r$256} - \texttt{C$_r$256} - \texttt{C$_r$256} - \texttt{C6}  
\end{itemize}
The overview for Attention PoseNet is given in Fig. \ref{fig:EndoSfM} b and the details of attention mechanism are introduced in next subsection. 

\subsubsection{Endo-SfMLearner Spatial Attention Block (ESAB)}
\label{ssec:ESAB}
 The intuition behind ESAB module in encoder layers is to guide pose network by emphasizing texture details and depth differences of pixels. On the contrary to feature-based and object-based attentions, spatial attention selects a specific region of the input image and features in that regions are processed by attention block. The ESAB mechanism is non-local convolutional process. For any given input $\mathbf{X} \in \mathcal{R}^{N\times64\times H \times W}$, our block operation can be overviewed as:

\begin{equation}
\mathbf{Z}=\mathit{f}(\mathbf{X}, \mathbf{X}^\top)\mathit{g}(\mathbf{X}),
\end{equation}
where $\mathit{f}$ stands for the pixelwise relations of input $\mathbf{X}$ between each pixel. The non-local operator extracts the relative weights of all positions on the feature maps.

In ESAB Block, we employ the dot product operation on max-pooled $\phi$ and $\theta$ convolution, which is activated by ReLU function:
\begin{equation}
    \mathbf{P} = \psi(\sigma_{relu}(\theta(\mathbf{X})\phi(\mathbf{X})^\top)),
\end{equation}
where $\sigma_{relu}$ is the ReLU activation function. The dot product, $\theta(\mathbf{X})\varphi(\mathbf{X})^\top$, gives a measurement for the input covariance, which can be defined as a degree of tendency between two feature maps at different channels.
We activate the $\psi$ convolution operation in $\mathit{softmax}$ function and perform a matrix multiplication between the $g$ and the output of $\mathit{softmax}$ function. Then, we convolve and upsample the result of multiplication with $\phi$ to extract the attention map $\mathbf{S}$. Finally, an element-wise sum operation in between attention map $\mathbf{S}$ and the input $\mathbf{X}$  generates the output $\mathbf{E} \in \mathbb{R}^{N\times64\times H\times W}$:
\begin{equation}
    \mathbf{S} = \phi (\sigma_{softmax}(\mathbf{P})g(\mathbf{X})),
\end{equation}
\begin{equation}
    \mathbf{F} = \mathbf{S} + \mathbf{X}, 
\end{equation}
where $\sigma_{softmax}$ denotes $\mathit{softmax}$ function. Short connection between the input $\mathbf{X}$ and the output $\mathbf{F}$ finalizes the block operations for the residual learning. The detailed flow diagram of block operations of ESAB module is given in Fig. \ref{fig:EndoSfM} d.

\subsubsection{Learning Objectives for Endo-SfMLearner}
\label{ssec:EndoSfM}

 Endo-SfMLearner is trained both in forward and backward directions with losses calculated in forward direction. We are using three loss functions to guide the network without labels; brightness-aware photometric loss, smoothness loss and geometry consistency loss. 
 
 Apart from well-known way of defining photometric loss, we are proposing affine brightness transformation between consecutive frames to deal with the problems stem from brightness constancy assumption of previous methods. First of all, the new reference image, $\hat{I}_i$, is synthesized via interpolating $I_{i+1}$. Previous methods calculate photometric loss directly comparing the synthesized image $\hat{I}_i$ with target image, $I_i$. However, the difference stem from illumination between consecutive frames might mislead the network. We propose to equate the brightness conditions between these two images as a robust way of supervising training phase. To the best of our knowledge, this is the first implementation of that approach for pose and depth estimation in literature. Moreover, quickly changing distance between organ tissue and camera results in scale inconsistency. We are using geometry consistency loss \citep{bian2019unsupervised} to cope with that problem. The overall objective of the system is to minimize the weighted sum of brightness-aware photometric loss $\mathcal{L}_{bp}^M$, smoothness loss $\mathcal{L}_s$ and geometry consistency loss $\mathcal{L}_{GC}$ which can be formulated as:
\begin{equation}
\label{eq:objective}
\mathcal{L} = \alpha\mathcal{L}_{bp}^M + \beta\mathcal{L}_s+\gamma\mathcal{L}_{GC}.
\end{equation}
where $\alpha$, $\beta$, and $\gamma$ are the weights for the related loss functions which are not necessarily adding up to one. 

The well-known photometric loss functions are based on the brightness constancy assumption which can be violated due to auto-exposure of the camera and fast illumination changes to which both L$_2$ and SSIM are no more invariant. To deal with that inconsistent illumination issue which is common in endoscopic image sequences, Endo-SfMLearner network predicts a brightness transformation parameter set which tries to align the brightness of input images during training on the fly and in a self-supervised manner. The evaluations demonstrate that the proposed brightness transformation significantly improves the pose and depth prediction accuracy.
The brightness-aware photometric loss formulation is given as follows: 

\begin{align}
\label{eq:photometricloss}
	\mathcal{L}_{bp} = \frac{1}{|P|} \mathlarger{\mathlarger{\sum}}_{p \in P}         &(\lambda_p \Vert{\mathbf{T}_b(\hat{I_i}(p))-I_{i}(p)}\Vert_2 \\ +&\lambda_s \frac{1-{SSIM}_{\mathbf{T}_b(\hat{I}_i),I_i}}{2} ) \nonumber
\end{align}

\begin{equation}
\label{eq:brightnessLoss2}
\mathbf{T}_b(\hat{I_i}) = \hat{I_{i}}^{a_{t'},b_{t'}} = a_{t\rightarrow t'} \hat{I_i} +c_{t\rightarrow t'} 
\end{equation}
where $\hat{I_i}$ stands for synthesized image by warping $I_{i+1}$, $\mathbf{T}_b$ is the brightness alignment function with affine transformation parameters $a_{t\rightarrow t'}$ and $c_{t\rightarrow t'}$, P stands for the successfully projected pixels from reference frame, $SSIM$ is the image dissimilarity loss. By making  use of contrast, luminance and  structure values of $\mathbf{T}_b(\hat{I_i})$  and $I_i$ image; SSIM targets to measure perceived image quality by human visual system and more sensitive to high frequency content such as textures and edges in regard of PSNR.  

Since the photometric loss is not sufficiently informative for the low-texture and homogeneous endoscopic images, we are also incorporating smoothness loss \citep{smoothnessLoss2018} which is calculated as a combination of predicted depth and input images for both reference and target frames.  
  
\begin{equation}
\label{eq:smoothnessLoss}
    \mathcal{L}_{s} =  \mathlarger{\mathlarger{\sum}}_{p \in P} (e^{-\nabla I_i(p)}.\nabla D_i(p))^2,
\end{equation}
where $\nabla$ is the first derivative along spatial directions. Thanks to the smoothness loss, Endo-SfMLearner is guided by edges in the predicted depth and input images. Finally, geometry consistency loss is integrated to our methodology. The main idea behind this loss is to confirm if $D_i$ provides the same scene under the transformation of $D_{i+1}$ by predicted relative poses $P_{i,i+1}$ . The difference between predicted depths, $D_{diff}$, can be calculated as: 
\begin{equation}
   \label{eq:depthDiff} 
   D_{diff}(p) = \frac{|D_{i+1}^i(p)-D_{i+1}'(p)|}{D_{i+1}^i(p)+D_{i+1}'(p)},
\end{equation}
where $D_{i+1}^i$ is the depth map of $I_{i+1}$ by warping $D_i$ via $P_{i,i+1}$ and $D_{i+1}'$ is the interpolated depth map from $D_{i+1}$. The geometry consistency loss will be defined as summation of this difference across all pixel coordinates after normalization with valid pixel counts:
\begin{equation}
   \label{eq:geometryconsistency} 
    \mathcal{L}_{GC} = \frac{1}{|P|} \mathlarger{\mathlarger{\sum}}_{p \in P} D_{diff}(p).
\end{equation}
This consistency constrain between consecutive depth maps paves the way for long trajectory estimation with higher accuracy, the reader is referred to see Fig. \ref{fig:EndoSfM} a. We also use depth inconsistency map results,  $D_{diff}$, to weight the $\mathcal{L}_{bp}$ with M as follows:
\begin{equation}
    \label{eqn:mask}
    M = 1 - D_{diff},
\end{equation}

\begin{equation}
    \mathcal{L}_{bp}^M = \frac{1}{|P|} \mathlarger{\mathlarger{\sum}}_{p \in P} (M(p).(\mathcal{L}_{bp}(p))).
    \label{eqn:maskWeightedLbp}
\end{equation}
Thanks to this operation, brightness-aware photometric loss is weighted with higher constant if the predicted and interpolated depth maps are inconsistent for each pixel. 
\begin{figure*}[h!]
\centering
\includegraphics[width=1.0\textwidth]{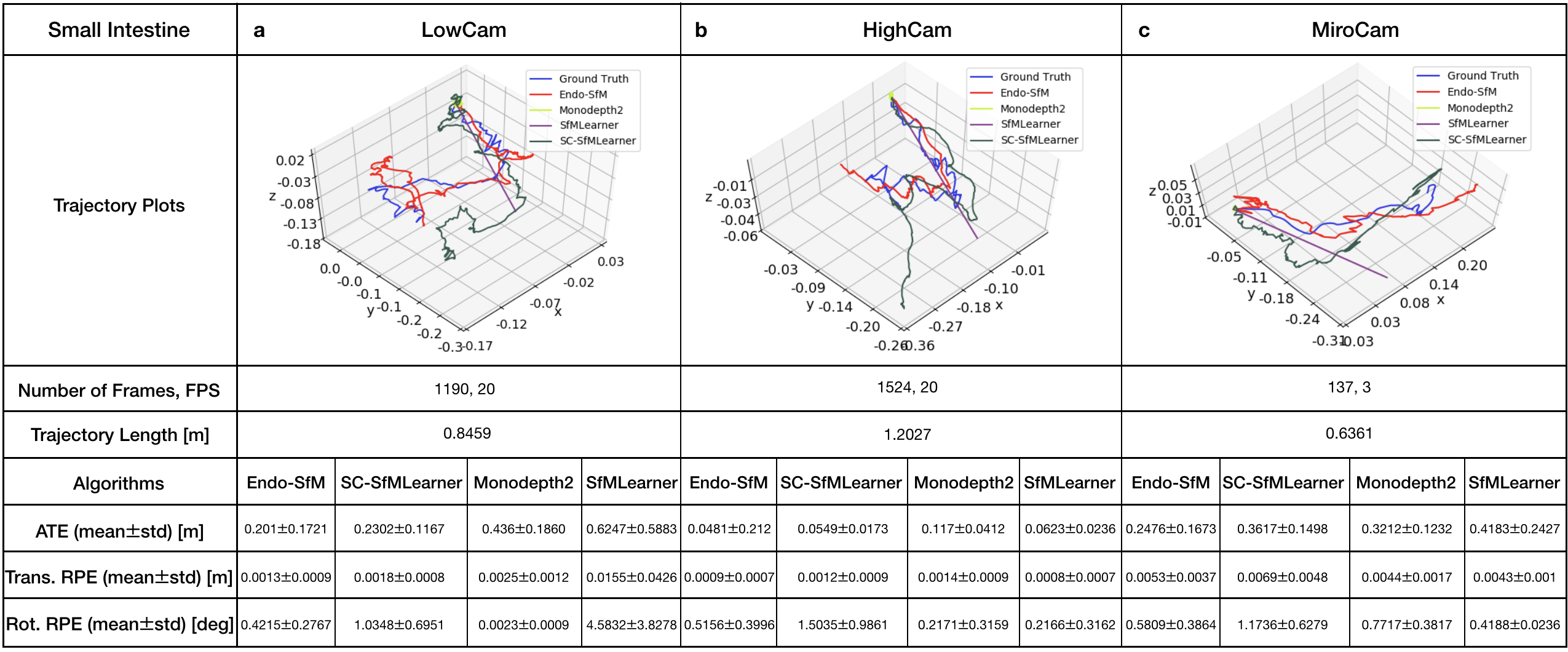}
\caption{\textbf{Pose estimations on real trajectories.} Endo-SfMLearner, SC-SfMLearner, Monodepth2, and SfMLearner trajectory estimations are benchmarked on ex-vivo EndoSLAM data. \textbf{a} The results for the first trajectory of small intestine recorded by LowCam. Throughout the all trajectory, SfMLearner relative pose estimations vary incredibly small which results in almost straightforward global pose estimation. We observed network firing problem for Monodepth2 even if we have repeated the tests on dataset with different frame-per-second rates.  However, SC-SfMLearner and Endo-SfMLearner exhibit more reasonable predictions thanks to the geometry consistency loss. The most challenging part of trajectories are sharp corners where position and orientation of camera change with high speed in small time intervals. At those points, Endo-SfMLearner yields performance with higher accuracy compared to SC-SfMLearner not just qualitatively but also quantitatively in terms of both translational and rotational errors.  \textbf{b} The results for first sub-trajectory of small intestine recorded by HighCam. It illustrates the case where SC-SfMLearner loose the scale consistency after sharp corner angle which is not the case for Endo-SfMLearner. Same problems observed for SfMLearner and Monodepth2 as in previous case. Endo-SfMLearner tracks loopy sections of the trajectories with sufficient precision up to 1000 frames by leaving small offset in between the ground truth. Apart from these, camera orientation estimations significantly improve and rotational relative pose error reduces almost three times compared to SC-SfMLearner which is the baseline state-of-the art method and achieves closest performance to ours. \textbf{c} The results for fourth trajectory of Colon-III recorded by MiroCam. On the contrary to the HighCam and LowCam, MiroCam exhibits fish-eye camera properties with high lens distortion. So, we have tested the reliability of Endo-SfMLearner against camera intrinsic properties.}
\label{fig:pos_comparisons}
\end{figure*}

\section{EndoSLAM Use-Case with Endo-SfMLearner}
\label{sec:analys}
 To illustrate the use-case of the EndoSLAM dataset, Endo-SfMLearner, our proposed learning-based structure-from-motion method was benchmarked for the pose and depth estimation tasks. Additionally, we have tested both dataset and EndoSfMLearner with a traditional fully dense 3D-reconstruction pipeline based on SIFT feature-matching and non-lambertian surface reconstruction where the detailed overview is given in Algorithm 1. Error metrics that were used to quantitatively assess the performance of the algorithms are introduced in the following subsections.

\subsection{Error Metrics}
Endo-SfMLearner pose estimation performance is tested based on three metrics: absolute trajectory error(ATE), translational relative pose error(trans RPE) and rotational relative pose error(rot RPE). The monocular depth estimation performance is evaluated in terms of Root Mean Square Error(RMSE). Finally, the 3D-reconstruction results are evaluated with surface reconstruction error. These error metrics are defined as follows based on  the estimated and ground truth trajectories represented by $\mathbf{P}_1, \ldots, \mathbf{P}_n \in \mathrm{SE(3)}$ and  $\mathbf{Q}_1, \ldots, \mathbf{Q}_n \in \mathrm{SE(3)}$, respectively, where the lower subscript is indexing frames and SE(3) is the Special Euclidean Group in three dimensions. 

\begin{algorithm}
\caption{3D Reconstruction and Evaluation Pipeline}

\begin{algorithmic}[1]
    \State Extract SIFT features between image pairs
    \State Find $k$-nearest neighbours for each feature using a  $k$-d tree
    
    \For{each image}
        \State (i) Select $m$ candidate matching images that have the  most number of corresponding feature points
        \State (ii) Find geometrically consistent feature matches using RANSAC to solve 
        for the homography between pairs of images. 
    \EndFor 
    \State Find connected components of image matches 
    \For{each connected component}
        \State (i) Perform bundle adjustment for connected components in image matches
        \State (ii) Render final stitched image using multi-band blending
    \EndFor 
\State Apply inpainting on the stitched image to suppress specularities
\State Reconstruct the surface using Tsai-Shah shape from shading method
\State Label a common line segment in ground truth data and reconstructed surface
\State Apply ICP algorithm using the common line as initialization
\State Compute iteratively the cloud-to-mesh distances to acquire RMSE

\end{algorithmic}
\end{algorithm}

\subsubsection{Absolute trajectory error (ATE)}
The ATE is a measure of global consistency between two trajectories, comparing absolute distances between ground truth and predicted poses at each point in time. Let the rigid body transformation ${\bf S}$ be the best (least-squares) alignment of the trajectories \citep{horn1987closed}.  
Then absolute trajectory error for the $i^{th}$ pose sample is calculated as follows:
\begin{equation}
    \textrm{ATE}_{i}=\lVert \textrm{trans}( {\bf Q}_{i}^{-1}{\bf SP}_{i} ) \lVert.
\end{equation}
The overall error throughout trajectory is defined by the root mean square of $\textrm{ATE}_{i}$.

\subsubsection{Relative Pose Error (RPE)}
\label{ssec:RPE} Relative pose error measures the difference in the change in pose over a fixed length $\Delta$ between two trajectories. Defining ${\bf E}_{i}(\Delta)=({\bf Q}_{i}^{-1}{\bf Q}_{i+\Delta})^{-1}({\bf P}_{i}^{-1}{\bf P}_{i+\Delta})$, the translational and rotational RPE are given by:
\begin{align}
    \textrm{Trans RPE}_i(\Delta) &= \lVert \textrm{trans}({\bf E}_i) \lVert, \\
    \textrm{Rot RPE}_i(\Delta) &= \angle (\textrm{rot}({\bf E}_i)),
\end{align}
where $\textrm{rot}({\bf E}_i)$ is the rotation matrix of ${\bf E}_i$ and $\angle(\cdot)$ is the positive angle of rotation. The errors are reported for $\Delta$ equals to 1.

\begin{table*}[!ht]
\caption{\textbf{Quantitative results of pose prediction for various organs and trajectories.} Endo-SfMLearner comparison with Endo-SfMLearner without attention block(Ew/oAtt), Endo-SfMLearner without brightness aware photometric loss integration(Ew/oBr), SC-SfMLearner, Monodepth2,  and SfMLearner. To test the algorithm robustness against tissue and trajectory differences, we performed tests on two separate trajectories from ex-vivo porcine stomach, colon and intestine. Absolute trajectory error (ATE), translational and rotational Relative Pose Error metrics are used as evaluation criteria. Moreover, for a better understanding of the camera specifications effect on pose estimation, we compared the results coming from high (HighCam) and low (LowCam) resolution camera for same trajectories. We observed considerable decrease in rotational errors for Endo-SfMLearner which proves the effectiveness of spatial attention block integrated to pose network encoder and brightness-aware photometric loss. Even though, most of the tests result in Endo-SfMLearner superiority, only for the third trajectory of Stomach-III from HighCam  SC-SfMLearner performed with higher accuracy in terms of ATE. Nevertheless, ablation studies do not provide sufficient cue to explain this improvement either stem from spatial attention block or brightness aware photometric loss.}
\resizebox{\textwidth}{!}{
\large 
\centering
\begin{tabular}{cc|cccc|cccc}

&Organ, Trajectory & \shortstack{Trajectory \\ Length \\\relax[m]}& \shortstack{ATE $\downarrow$\\(mean$\pm$ std) \\\relax [m]}& \shortstack{Trans. RPE $\downarrow$ \\ (mean$\pm$ std) \\\relax[m]} & \shortstack{Rot. RPE $\downarrow$ \\ (mean$\pm$ std)\\\relax [deg]} & \shortstack{Trajectory \\ Length \\\relax[m]} & \shortstack{ATE $\downarrow$\\(mean$\pm$ std)\\\relax [m]}& \shortstack{Trans RPE  $\downarrow$\\ (mean$\pm$ std) \\\relax[m]} & \shortstack{Rot RPE $\downarrow$ \\ (mean$\pm$ std)\\\relax [deg]}\\
\midrule\midrule[.1em]
 & &\multicolumn{4}{c}{HighCam} & \multicolumn{4}{c}{LowCam} \\ \cmidrule{1-10}

\parbox[c]{2mm}{\multirow{7}{*}{\rotatebox[origin=c]{90}{EndoSfM}}}
&Colon-IV,Traj-I & 0.4286 & 0.0878$\pm$ 0.0549 & 0.0009$\pm$ 0.0027 & 0.488$\pm$ 0.3217
&0.6785 
& 0.1046$\pm$ 0.0343& 0.0011$\pm$ 0.006 & 0.4666$\pm$ 1.3792 \\
&Colon-IV,Traj-V & 1.2547 &  0.1731$\pm$ 0.1179 & 0.0014$\pm$ 0.002 & 0.2552$\pm$ 0.417
&1.1699 & 
0.1771$\pm$ 0.1177 & 0.0012$\pm$ 0.002 & 0.1493$\pm$ 0.2321\\

&Intestine,Traj-IV & 1.0557 & 0.0812$\pm$ 0.0152 & 0.0010 $\pm$ 0.0013 & 0.173$\pm$ 0.1942 
&0.8265 &
0.0558$\pm$ 0.0356 & 0.0011$\pm$ 0.0008 & 0.404 $\pm$ 0.5052\\ 
&Stomach-I,Traj-I & 1.4344 & 0.1183$\pm$ 0.1062 & 0.0013$\pm$ 0.0028 & 0.5988$\pm$ 0.8185
& 0.8406 &
0.1732$\pm$ 0.116 & 0.0021$\pm$ 0.0034 & 0.8424$\pm$ 1.0788\\
&Stomach-III,Traj-III & 0.8908 & 0.1177$\pm$ 0.0543 & 0.0013$\pm$ 0.0033 & 0.5543$\pm$ 0.928
& 0.9714 & 
0.1014$\pm$ 0.0491 &0.0011$\pm$ 0.0007 & 0.6705$\pm$ 0.3817\\

\midrule
\parbox[c]{2mm}{\multirow{7}{*}{\rotatebox[origin=c]{90}{Ew/oAtt}}}
&Colon-IV,Traj-I & 0.4286 & 0.0894$\pm$ 0.0274 & 0.0010$\pm$ 0.0029 & 0.3502$\pm$ 0.2621 
&0.6785 &
0.1548$\pm$ 0.0591 & 0.0010$\pm$ 0.3679 & 1.3613$\pm$ 1.5908 \\ 

&Colon-IV,Traj-V & 1.2547 & 0.1855 $\pm$ 0.0494 & 0.0014$\pm$ 0.0022 & 0.4569$\pm$ 0.5734
&1.1699 & 
0.1628$\pm$ 0.0375 & 0.0014$\pm$ 0.003 & 0.4168$\pm$ 0.3149\\

&Intestine,Traj-IV & 1.0557 & 0.1055$\pm$ 0.0379 & 0.0011$\pm$ 0.0012 & 0.3343$\pm$ 0.2653 
&0.8265 &
0.0691$\pm$ 0.0305 & 0.001$\pm$ 0.0009 & 0.654 $\pm$ 0.6042\\ 

&Stomach-I,Traj-I & 1.4344 & 0.1889 $\pm$ 0.0497 & 0.0015$\pm$ 0.0038 & 0.893$\pm$ 0.915
& 0.8406 &
0.1968$\pm$ 0.1417 & 0.0025$\pm$ 0.0037 & 1.1823$\pm$ 1.2112\\ 

&Stomach-III,Traj-III & 0.8908 & 0.1362 $\pm$ 0.068 & 0.0016$\pm$ 0.0032 & 0.8244 $\pm$ 1.0127 
& 0.9714 & 
0.1204$\pm$ 0.0418 & 0.0010$\pm$ 0.0009 & 1.0907$\pm$ 0.5634\\

\midrule 
\parbox[c]{2mm}{\multirow{7}{*}{\rotatebox[origin=c]{90}{Ew/oBr}}}
&Colon-IV,Traj-I & 0.4286 & 0.1328$\pm$ 0.0431 & 0.0010$\pm$ 0.0026 & 0.7198$\pm$ 0.4764
&0.6785 &
0.1402 $\pm$ 0.0671 & 0.0010$\pm$ 0.0060 & 0.7257$\pm$ 1.424 \\
&Colon-IV,Traj-V & 1.2547 &  0.1898$\pm$ 0.0709 & 0.0015$\pm$ 0.002 & 0.929$\pm$ 0.7525
&1.1699 & 
0.1503$\pm$ 0.0433 & 0.0013$\pm$ 0.002 & 0.8989$\pm$ 0.6199\\

&Intestine,Traj-IV & 1.0557 & 0.1467$\pm$ 0.0848 & 0.002 $\pm$ 0.0010 & 0.6607$\pm$ 0.3884 
&0.8265 &
0.1241 $\pm$ 0.0436 & 0.0009$\pm$ 0.0008 & 1.106 $\pm$ 0.8081\\ 
&Stomach-I,Traj-I & 1.4344 & 0.1963 $\pm$ 0.0478 & 0.002 $\pm$ 0.0032 & 0.6899 $\pm$ 1.0401 
& 0.8406 &
0.1923$\pm$ 0.118 & 0.0023$\pm$ 0.0032 & 0.9215$\pm$ 1.1728\\
&Stomach-III,Traj-III & 0.8908 & 0.1277$\pm$ 0.0805 & 0.0014$\pm$ 0.0033 & 0.3933$\pm$ 0.9258
& 0.9714 & 
0.1101 $\pm$ 0.0257 & 0.0010$\pm$ 0.0006 & 0.439 $\pm$ 0.2672 \\

\midrule
\parbox[c]{2mm}{\multirow{7}{*}{\rotatebox[origin=c]{90}{SC-SfM}}}
&Colon-IV,Traj-I & 0.4286 & 0.1545$\pm$ 0.0441 & 0.0014$\pm$ 0.0028 & 1.3532$\pm$ 0.8541 
&0.6785 &
0.1898$\pm$ 0.0718 & 0.0015$\pm$ 0.0060 & 1.6388$\pm$ 1.5908 \\ 

&Colon-IV,Traj-V & 1.2547 & 0.2054$\pm$ 0.1734 & 0.0024$\pm$ 0.0029 & 1.2452$\pm$ 0.965
&1.1699 & 
0.1667$\pm$ 0.1263 & 0.0021$\pm$ 0.003 & 1.2188$\pm$ 0.7715\\

&Intestine,Traj-IV & 1.0557 & 0.1247$\pm$ 0.1327 & 0.0015$\pm$ 0.0009 & 0.9257$\pm$ 0.584 
&0.8265 &
0.0908$\pm$ 0.0819 & 0.0016$\pm$ 0.0009 & 0.8989 $\pm$ 0.7854\\ 

&Stomach-I,Traj-I & 1.4344 & 0.2325$\pm$ 0.127 & 0.002$\pm$ 0.0038 & 1.2937$\pm$ 1.2484
& 0.8406 &
0.191$\pm$ 0.1399 & 0.0028$\pm$ 0.0033 & 2.1322$\pm$ 1.2601\\ 

&Stomach-III,Traj-III & 0.8908 & 0.0898 $\pm$ 0.035 & 0.0016$\pm$ 0.0033 & 1.3071$\pm$ 1.3187 
& 0.9714 & 
0.1927$\pm$ 0.0561 & 0.0012$\pm$ 0.0007 & 2.041$\pm$ 0.8391\\

\midrule
\parbox[c]{2mm}{\multirow{7}{*}{\rotatebox[origin=c]{90}{Mono2}}}
&Colon-IV,Traj-I & 0.4286 & 0.1071$\pm$ 0.0756 & 0.0012$\pm$ 0.0028 & 0.3115$\pm$ 0.268
&0.6785 
& 0.215$\pm$ 0.1084 & 0.0009$\pm$ 0.006 & 0.1679$\pm$ 1.378 \\
&Colon-IV,Traj-V & 1.2547 & 0.1872$\pm$ 0.1404 & 0.0016$\pm$ 0.002 & 0.1607$\pm$ 0.4226
&1.1699 & 
0.2158$\pm$ 0.1466 & 0.0018$\pm$ 0.002 & 0.3921$\pm$ 0.3362\\

&Intestine,Traj-IV & 1.0557 & 0.1507$\pm$ 0.1165 & 0.009 $\pm$ 0.0013 & 0.1092$\pm$ 0.1812 
&0.8265 &
0.1431$\pm$ 0.132 & 0.0014$\pm$ 0.001 & 0.3128 $\pm$ 0.5288\\ 
&Stomach-I,Traj-I & 1.4344 & 0.2878 $\pm$ 0.2293 & 0.0029$\pm$ 0.0038 & 0.298$\pm$ 0.7968
& 0.8406 &
0.2033$\pm$ 0.0971 & 0.0019$\pm$ 0.0011 & 0.5296$\pm$ 0.3642\\ 
&Stomach-III,Traj-III & 0.8908 & 0.5841$\pm$ 0.2742 & 0.0022$\pm$ 0.0033 & 0.8178$\pm$ 0.9059 
& 0.9714 & 
0.3876$\pm$ 0.2322 & 0.0032$\pm$ 0.0017 & 0.7345$\pm$ 0.8349\\

\midrule
\parbox[c]{2mm}{\multirow{7}{*}{\rotatebox[origin=c]{90}{SfM}}}
&Colon-IV,Traj-I & 0.4286 & 0.1584$\pm$0.1064 & 0.0043$\pm$ 0.0042 & 2.6624$\pm$ 1.6822 
&0.6785 
& 0.1946$\pm$ 0.1708 & 0.0037$\pm$ 0.0092 & 2.0718$\pm$ 2.3018 \\
&Colon-IV,Traj-V & 1.2547 & 0.5849$\pm$ 0.5201 & 0.0092$\pm$ 0.0175 & 4.4083$\pm$ 4.6309 
&1.1699 &
0.2094$\pm$ 0.1613 & 0.005$\pm$ 0.0041 & 3.1999$\pm$ 1.8304\\

&Intestine,Traj-IV & 1.0557 & 0.2119$\pm$ 0.2022 & 0.0083$\pm$ 0.016 & 3.9877$\pm$ 5.2134
&0.8265 & 
0.2387$\pm$ 0.1675 & 0.0048$\pm$ 0.005 & 2.7019$\pm$ 2.189\\
&Stomach-I,Traj-I & 1.4344 & 0.1741$\pm$ 0.0744 & 0.0012$\pm$ 0.0038 & 0.7249$\pm$ 0.7904
& 0.8406 &
0.2226$\pm$ 0.0989 & 0.007$\pm$ 0.005 & 4.1709$\pm$ 2.3479\\ 
&Stomach-III,Traj-III & 0.8908 & 0.3086$\pm$ 0.1774 & 0.0018$\pm$ 0.0035 & 0.6137$\pm$ 0.996 
& 0.9714 & 
0.1711$\pm$ 0.0548 & 0.0012$\pm$ 0.0008 & 0.802$\pm$ 0.4236\\

\midrule[.1em]
\end{tabular}
}
\label{tab:trajResults}
\end{table*}

\subsubsection{Surface Reconstruction Error}
We use the methodology propounded by \citep{handa2014benchmark} in order to evaluate the surface reconstruction quality. As the first step, one line segment is manually identified between the reconstructed and ground truth 3D maps. The match points are used to coarsely align both maps. This coarse alignment is used as an initialization for the iterative closest point (ICP) algorithm. ICP iteratively aligns both maps until a termination criteria of 0.001 cm deviation in RMSE is reached.

\subsection{Pose Estimation with Endo-SfMLearner}
\label{ssec:SLAM}
All methods including Endo-SfMLearner are trained with the same data and parameter set for the sake of fairness and unbiased results. The training and validation dataset consist of 2,039 and 509 colon images generated in the Unity simulation environment, respectively. We train the all networks in 200 epochs with randomly shuffled batches each size of 4 images, optimize by ADAM with initial learning rate $10^{-4}$ and validate after each epoch. According to the tests in terms of ATE, trans RPE, and rot RPE on the data recorded via the HighCam and LowCam, Endo-SfMLearner achieves the state-of-the-art for most of the cases. The results in Table~\ref{tab:trajResults} show clear advantage of ESAB block integration and brightness-aware photometric loss. In the majority of Stomach-III results for both HighCam and LowCam, all models fail to follow trajectory with sufficient accuracy. However, the predicted trajectories aligned with ground truth for Endo-SfMLeaner in general are much better compared to other models.

Both quantitative and qualitative pose estimation results on sample trajectories of HighCam, LowCam and MiroCam are given in Fig.~\ref{fig:pos_comparisons}. Under the above mentioned training conditions, Monodepth2 and SfMLearner face with fatal failure on endoscopic videos. Even if SC-SfMLearner exhibits closest performance to our method in terms of absolute trajectory errors, we observe improvement specially on the rotational movement estimations which is reflected on rotational relative pose errors. A similar observation is also made on Unity trajectories, see  Fig.~\ref{fig:unityposeestimation}. Since the rotations cannot be changed frequently and easily while recording clear images in Unity environment, the trajectories are close to the straight lines which result in higher accuracy for all methods. It is seen that the Endo-SfMLearner outputs generally follow the shape of the ground truth, specifically it catches rotations more consistently which is the main reason for the decrease in rotational relative pose error.

For more comprehensive evaluations of results in terms of camera motions, descriptive analysis of the camera speeds and accelerations are given in Fig.~\ref{fig:histo}, Table~\ref{tab:trajSpeedAccHigh}  and Table~\ref{tab:trajSpeedAccLow}. Since the robot motions are highly effective on image quality, we expect decrease in the pose estimation accuracy for the trajectories of Stomach-III which have highest mean speed and acceleration. The fact pave the way for the difficulty in alignment of those trajectories and also stitching of those frames for 3D reconstruction.

\subsection{Depth Estimation with Endo-SfMLearner}
\label{ssec:EndoSFMdepth}
In this subsection, we analyse the monocular depth estimation performance of Endo-SfMLearner quantitatively on synthetically generated data coming from EndoSLAM dataset. 

\begin{figure*}[!ht]
    \centering
    \includegraphics[width=2.1\columnwidth]{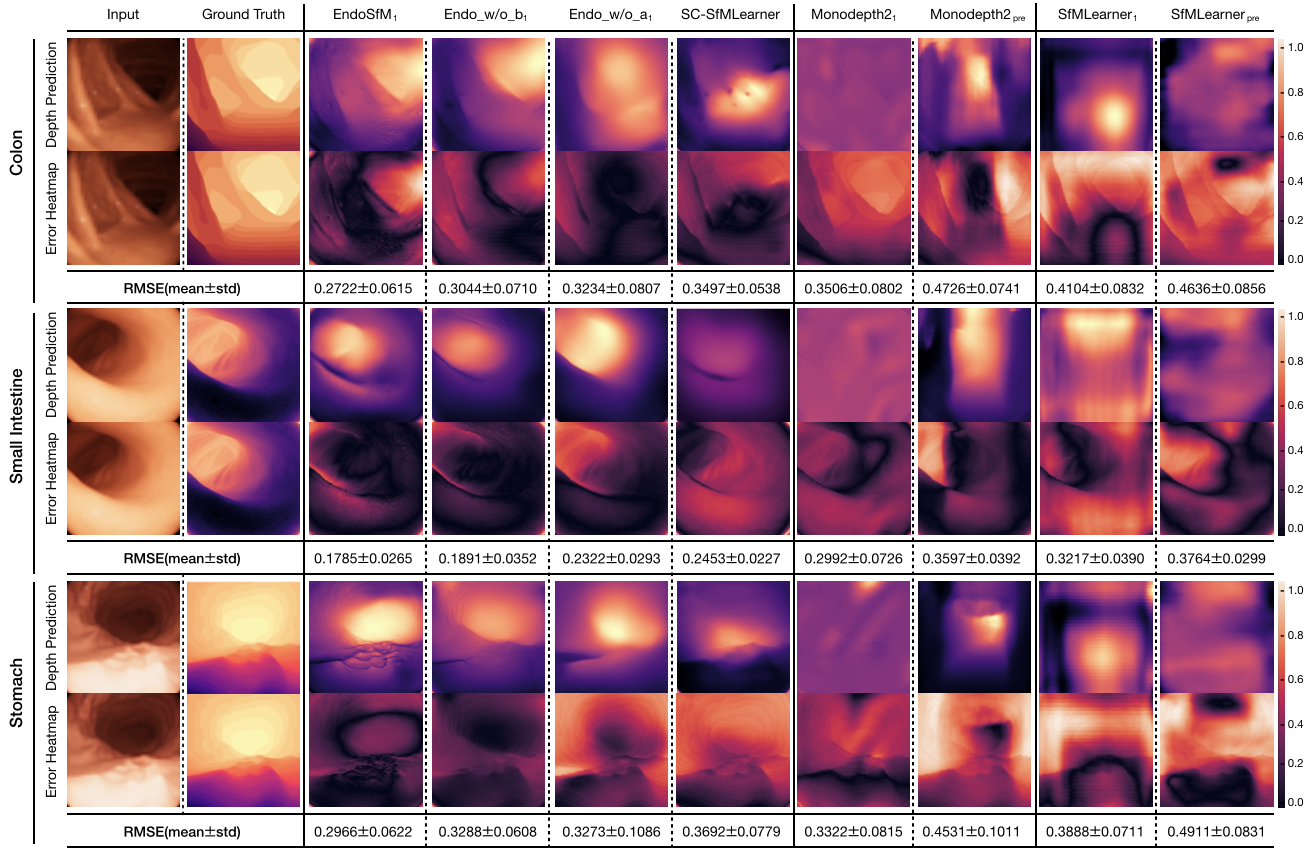}
    \caption{\textbf{Quantitative Depth Evaluations.} The original input image, depth ground truth, predicted depth maps and error heatmaps by Endo-SfMLearner, Endo-SfMLearner without brightness loss integration(Endo$\textunderscore$w/o$\textunderscore$b$_{1}$), Endo-SfMLeaner without attention integration(Endo$\textunderscore$w/o$\textunderscore$a$_{1}$), SC-SfMLearner(Endo-SfMLearner without loss and block operation), Monodepth2, published pretrained Monodepth2(Monodepth2$_{pre}$), SfMLearner and published pretrained SfMLearner(SfMLearner$_{pre}$) are shown from left to right, respectively. We benchmark the algorithms quantitatively on the synthetically generated images acquired with the camera whose properties are equivalent to the MiroCam. Even if the models subscripted by "1" are trained with the same data and parameter set, Endo-SfMLearner and SC-SfMLearner which are guided by geometry consistency loss show considerably superior performance to the rest of the methods. In particular, Endo-SfMLearner is able to estimate the relatively far regions more accurately than the remaining ones, although it is optimized for the images obtained by shallow Depth of Field cameras. Besides, its predictions conform with camera light burst and small depth alterations which result in least RMSE errors for all organs that is also proving the cross-organ adaptability of the method. By comparing the Endo-SfMLearner,  Endo$\textunderscore$w/o$\textunderscore$b$_{1}$ and Endo$\textunderscore$w/o$\textunderscore$a$_{1}$, one can deduce that the biggest advantage of ESAB block in PoseNet provided to the DispNet is increasing texture awareness whereas brightness-aware photometric loss focus the network to the light variations throughout the pixels. Their collaboration significantly improve the performance which is supported by decreasing RMSE values. The published
   pre-trained models are trained with Kitty dataset genarally consist of images whose upper part representing distant sky points, right
  and left edges are closer points representing flats or moving cars. This fact causes biased depth estimation especially for Monodepth2$_{pre}$, on
  endoscopic images from all organs.}
    \label{fig:depth}
\end{figure*}

Since EndoSLAM dataset also provides pixelwise depth ground truth for synthetically generated endoscopic frames, we show that Endo-SfMLearner quantitatively outperforms the benchmarked monocular depth estimation methods as given in Fig. \ref{fig:depth}. The results are evaluated in terms of root mean square error(RMSE) on 1,548 stomach, 1,257 small intestine and 1,062 colon frames. Even if the training and validation dataset consist of synthetic colon frames, Endo-SfMLearner depicts high performance on stomach and small intestine with 0.2966 and 0.1785 mean RMSE. The heatmaps are also indicating that the errors significantly decrease for the pixels representing regions far from 14mm.

\subsubsection{Ablation Studies for Spatial Attention Block}
\label{ssec:ablation}
In order to increase the pose and depth network sensitivity for the edge and texture details, we have integrated attention block in between ReLU and max pooling operations in PoseNet encoder. By this attention mechanism, we are expecting to preserve low and high-frequency information from the input endoscopic images by exploiting the feature-channel inter-dependencies. In this subsection, we specifically investigate the following cases:
\begin{itemize}
    \item EndoSfMLearner with brightness-aware photometric loss and ESAB,
    \item EndoSfMLearner with ESAB and without brightness-aware photometric loss,
    \item EndoSfMLearner without ESAB and with 
    brightness-aware photometric loss,
    \item EndoSfMLearner without ESAB and without
    brightness -aware photometric loss(SC-SfMLearner).
\end{itemize}
\begin{figure*}[!ht]
    \centering
    \includegraphics[width=2\columnwidth]{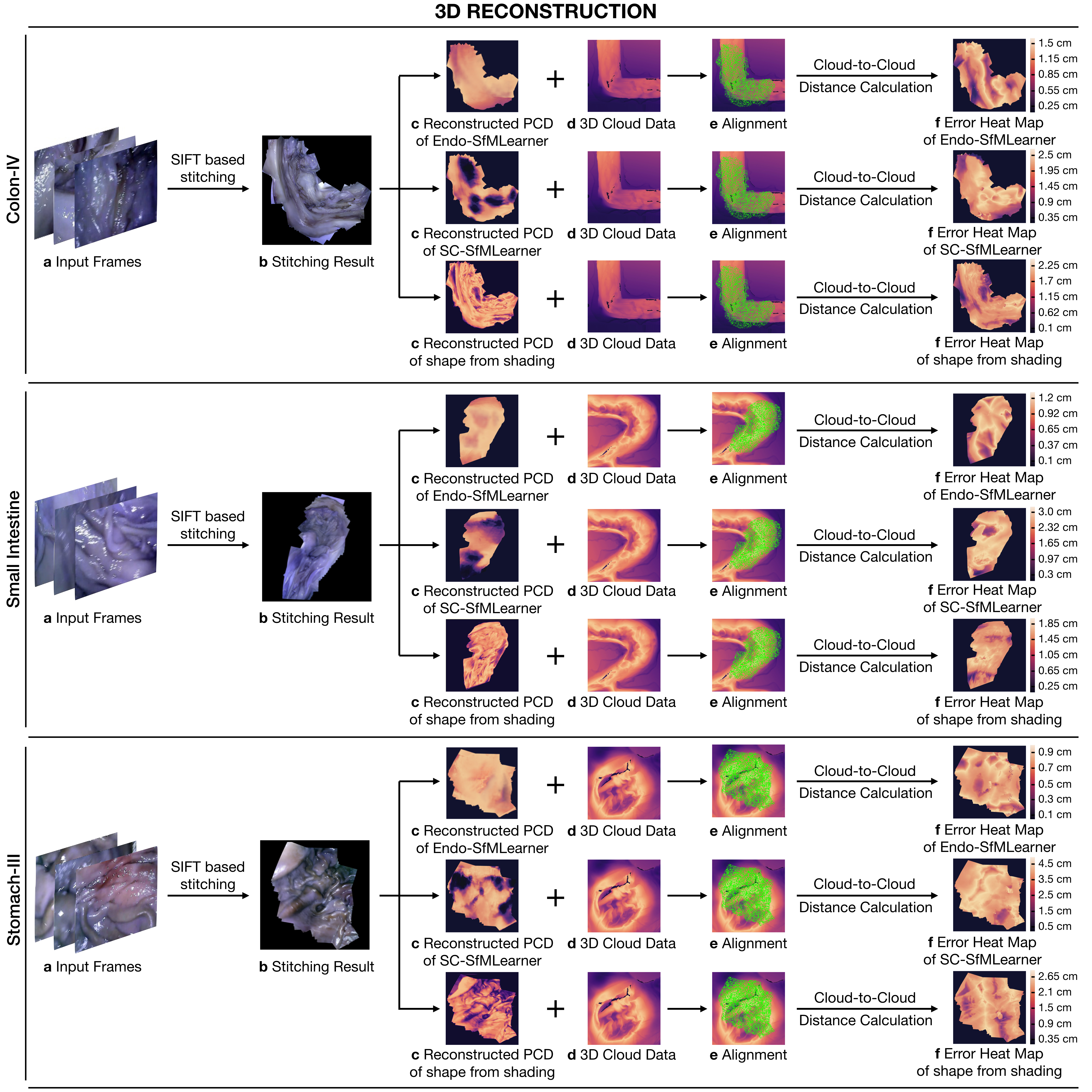}
    \caption{\textbf{3D Map Reconstruction and Evaluation Pipeline.} {\textbf{a}} Input image sequences from Colon-IV, Small Intestine and Stomach-III trajectories which are downsampled to 4 fps. The frames are given as input to Scale Invariant Feature Transform (SIFT), separately. {\textbf{b}} The final stitched image which is formed by aligning and blending all input frames. Specularities are suppressed using inpainting function of OpenCV. {\textbf{c}} Depth maps for inpainted images which are predicted using Endo-SfMLearner, SC-SfMLearner and shape from shading. {\textbf{d}} 3D scanner point cloud data for each organ in ply-format. {\textbf{e}} The matched area between reference and aligned cloud points by emphasizing in green colour. The aligned regions are chosen as same for all compared groups for the sake of fairness. Iterative Closest Point(ICP) was used to align the ground truth data and reconstructed surface after manually labelling a common line segment. {\textbf{f}} The cloud mesh distances in the form of heatmap with the bar displaying the root mean square error in cm. The RMSE values of Colon-IV, 0.51 cm, 0.86 cm and 0.65 cm for Endo-SfMLearner, SC-SfMLearner and shape from shading, respectively. The RMSE values of Small Intestine are 0.40 cm, 1.02 cm and 0.54 cm for Endo-SfMLearner, SC-SfMLearner and shape from shading, respectively. The RMSE values of  Stomach-III are 0.41 cm, 1.37 cm and 0.73 cm for  Endo-SfMLearner, SC-SfMLearner and shape from shading, respectively. For all organs, we sight the superiority of the Endo-SfMLearner over both SC-SfMLearner and shape from shading. Since the training and validation dataset of SC-SfMLearner consist of colon frames, the RMSE values for colon are smaller than the other organs. However, even if the Endo-SfMLearner has the same training and validation dataset, it exhibits highly effective performance on stitched stomach and intestine images in comparison with the remaining methods. }
    \label{fig:lowCam_effects_stitch}
\end{figure*}
The results for the pose tracking given in Table \ref{tab:trajResults} reveal the usefulness and effectiveness of the module. Although the attention module is only inserted in PoseNet, simultaneous training of networks causes the improvement in depth estimation which is depicted in Fig. \ref{fig:depth}. As seen from quantitative ablation analysis, attention module makes Endo-SfMLearner more responsive for depth alterations on the synthetically generated images from colon, intestine and stomach.   Even for the stomach and small intestine that is not included in training phase, Endo-SfMLearner achieves acceptable RMSE values which is the indicator of its persistent effort to be adaptable for texture differences.

\subsection{3D Reconstruction Pipeline}
\label{ssec:3drec}
3D reconstruction approach used in this work is depicted in Algorithm1. The steps of the procedure can be summarized under four fundamental techniques which are Otsu threshold-based reflection detection, OPENCV inpainting-based reflection suppression, feature matching and tracking based image stitching and non-lambertion surface reconstruction. To establish feature point correspondences between frames,  SIFT feature matching and RANSAC based pair elimination are employed \citep{AutoStitch2007}. Then, the depth map is estimated using Tsai-Shah shape from shading approach.

This surface reconstruction method applies a discrete approximation of the gradients and then utilize the linear approximation of the reflectance function in terms of the depth directly. For further details of Tsai-Shah method, the reader is referred to the original paper \citep{ping1994shape}.
Fig. \ref{fig:lowCam_effects_stitch} demonstrates the steps of the approach and output maps aligned with the ground truth scanned data. Using that pipeline, RMSEs of 0.65 cm, 0.54 cm and 0.73 cm are obtained for Colon-IV, Small Intestine and Stomach-III trajectories, respectively. Besides, we use the depth estimations of EndoSfMLearner and SC-SfMLearner on the stitched images and compared with the rule-based method. As a result, we get lower RMSE values on the aligned images for Endo-SfMLearner, see Fig. \ref{fig:lowCam_effects_stitch}.

\section{Discussion and Future Works}
\label{sec:discussion}
In this paper, we introduce a novel endoscopic SLAM dataset that contains both capsule and standard endoscope camera images with 6D ground truth pose and high precision scanned 3D maps of the explored GI organs. Four different cameras were employed in total to collect data from eight ex-vivo porcine GI-tract organs each from different animal instances. Various additional post processing effects such as fish eye distortions, Gaussian blur, downsampling and vignetting can be applied as optional to diversify and enrich the dataset. In addition to the EndoSLAM dataset, Endo-SfMLearner is proposed as a monocular pose and depth estimation method based on spatial attention mechanisms and brightness-aware hybrid loss. Although Endo-SfMLearner is specifically developed and optimized for endoscopic type of images, it also holds great promise for laparoscopy images due to similar texture characteristics. Our future work will focus on generalizing the EndoSLAM dataset concept to other visualization techniques and create datasets with various other imaging modalities. Furthermore, we aim to examine and improve the data adaptability of the Endo-SfMLearner and address these issues as next steps. Last but not least, we plan to investigate the combination of Endo-SfmLearner with segmentation, abnormality detection and classification tasks in the concept of multi-task and meta-learning to enhance the performance of state-of-the-art methods.

\section*{Acknowledgment}
\label{sec:acknowledgement}
Mehmet Turan, Kutsev~Bengisu~Ozyoruk, 
 Guliz Irem Gokceler, Gulfize Coskun, and Kagan Incetan are especially grateful to the Scientific and Technological Research Council of Turkey (TUBITAK) for International Fellowship for Outstanding Researchers. We would like to express deep gratitude to Abdullhamid Obeid and Ebru Sagiroglu for their valuable support during experiments.

\bibliography{references}
\clearpage

\onecolumn

\appendix
\counterwithin{figure}{section}
\counterwithin{table}{section}
\section{Equipment}
\begin{figure*}[!ht]

{\includegraphics[width = \columnwidth]{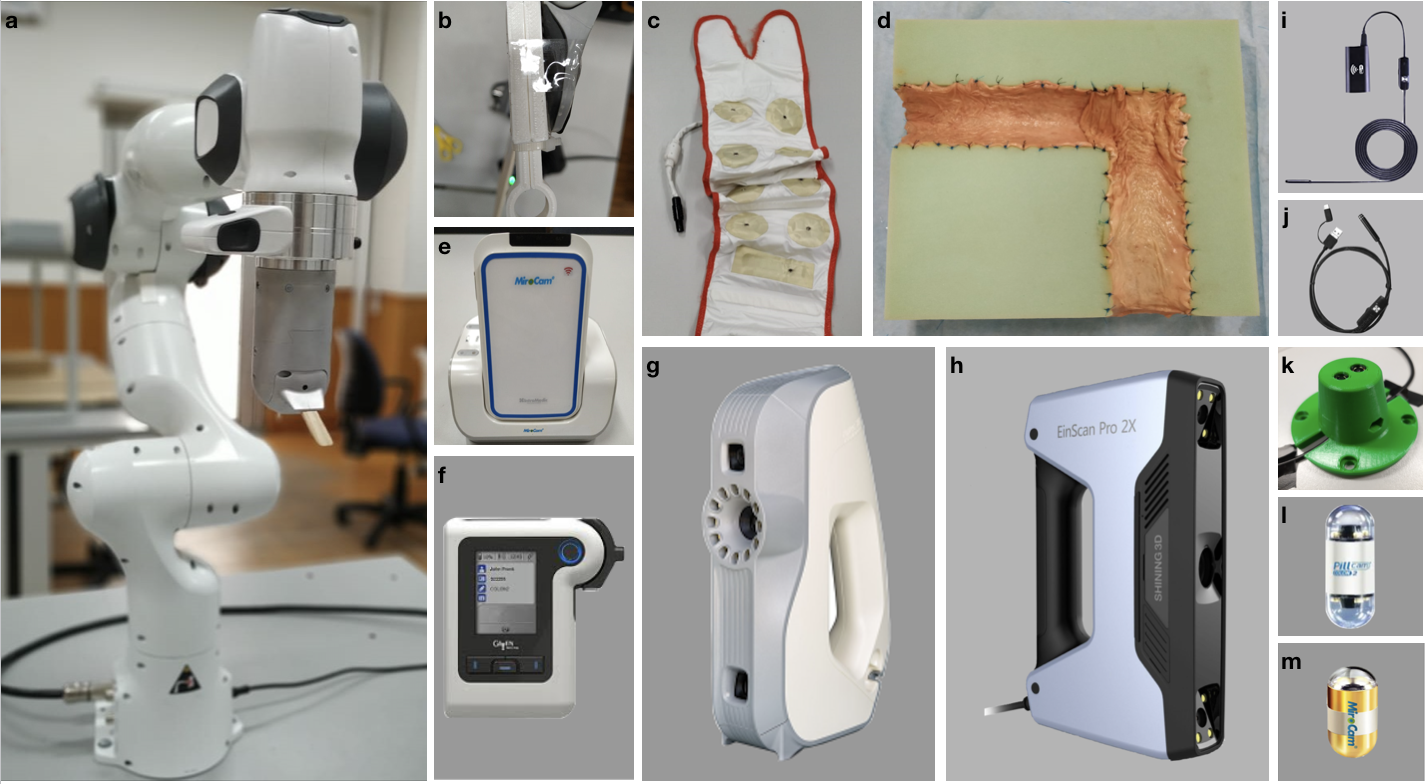}
} 
\caption{
	\textbf{Equipment.} The overall equipment for dataset generation. 
	\textbf{a} Franka Emika Panda: motion control device for cameras.
	\textbf{b} Capsule Holder: two-piece holder as a kit between the WCE cameras and the robotic arm.
	\textbf{c} MiroCam\textsuperscript{\textregistered} Data Belt 
	\textbf{d} Real Porcine Colon: sewn onto an 'L' shaped semi-cylindrical scaffold in high-density foam.
	\textbf{e} MiroCam\textsuperscript{\textregistered}  MR1100 receiver: Digital video grabber for the conversion of analog data into digital and output to the computer.
	\textbf{f} PillCam\textsuperscript{\textregistered} recorder
	\textbf{g} Artec Eva: 3D scanner used to generate ground truth - ply file. 
	\textbf{h} EinScan Pro 2X: 3D scanner used to generate ground truth - .ply, .obj, .stl and .ASC file.
	\textbf{i} Wireless Endoscope Camera (YPC-HD720P): high resolution - 1280×720 and HD640×480. 
	\textbf{j} Endoscope 3 in 1 Camera: low resolution - 640×480.
    \textbf{k} Camera Holder: specially designed one-piece holder for the stabilization of the high and low resolution endoscope to the robotic arm.
    \textbf{l} PillCam\textsuperscript{TM} COLON2: WCE double tip camera.
    \textbf{m} MiroCam\textsuperscript{\textregistered} Regular MC1000-W: WCE camera. 
}
\label{fig:equipment}
\end{figure*}

\section{Dataset Survey}
\label{sec:survey}
\begin{itemize}
\item The KID Dataset
is organized by The Medical Decision Support Systems (MDSS) research group of the University of Thessaly. The dataset is divided into two annotated sections. The first section has a total of 77 wireless capsule endoscopy (WCE) images acquired using MiroCam\textregistered\ (IntroMedic Co, Seoul, Korea) capsules and has some types of abnormalities such as angioectasias, apthae, chylous cysts and polypoid lesions. The second part consists of 2,371 MiroCam\textregistered\ WCE. This dataset not only includes small bowel lesions such as polypoid, vascular and inflammatory lesions but also images from healthy esophagus, stomach, small bowel and colon
Given Imaging Atlas Dataset consists of 20 second video clips recorded using PillCam capsules with  a resolution of 576x576 pixels. In this database, 117 WCE video clips have been acquired from the small bowel, 5 from esophagus and 13 from the colon \citep{spyrou2013video}. 

\item The Kvasir dataset was collected via standard endoscopic equipments at Vestre Viken (VV) Health Trust in Norway. The initial dataset consists of 4,000 images with eight classes namely Z-line, pylorus, cecum, esophagitis, polyps, ulcerative colitis, dyed and lifted polyps and dyed resection margins of images, each represented with 500 images. All images are annotated and verified by experienced endoscopists \citep{Kvasir2017}. Later, the dataset extended to 8,000 images with the same eight classes \citep{HyperKvasir}. The Kvasir-SEG Dataset is an extension of the Kvasir dataset which is used for polyp segmentation. It comprises 1000 polyp images and their corresponding ground truth from the second version of the Kvasir dataset \citep{kvasir_seg}.
\item The Hyper-Kvasir dataset is the largest online available dataset related to the gastrointestinal tract, containing 110,079 images (10,662 labeled and 99,417 unlabeled images) and 373 videos, making a total of 1.17 million frames. The entire dataset was collected in gastro- and colonoscopy examinations in Norway and 10,662 images are labeled for 23 classes by practitioners.  \citep{HyperKvasir}. 
\item The NBI-InfFrames dataset includes Narrow-band imaging(NBI) endoscopy which is commonly used as a diagnostic procedure to examine the back of throat, glottis, vocal cords and the larynx. To generate this in vivo dataset,  18 different patients affected by laryngeal spinocellular carcinoma (diagnosed after histopathological examination) were involved. It consists of 180 informative (I), 180 blurred (B), 180 with saliva or specular reflections (S) and 180 underexposed (U) frames with a total number of 720 video frames \citep{NBIInFrameDataset}.
\item The EndoAbS(Endoscopic Abdominal Stereo Images) Dataset consists of 120 sub-datasets of endoscopic stereo images of abdominal organs (e.g., liver, kidney, spleen) with corresponding ground truth acquired via laser scanner. In order to create variations in the dataset, frames have been recorded under 3 different lighting conditions, presence of smoke and 2 different distances from endoscope to phantom ($\sim$ 5 cm and $\sim$ 10 cm). The main purpose of generating this dataset was to validate 3D reconstruction algorithms for the computer assisted surgery community \citep{endoabsdataset}.
\item CVC-ColonDB is a database of annotated video sequences consisting of 15 short colonoscopy sequences, where one polyp has been shown in each sequence. There are 1,200 different images containing original images, polyp masks, non-informative image masks and contour of polyp masks. It can be used for assessment of polyp detection \citep{CVCColonDB}.
\item MICCAI 2015 Endoscopic Vision Challenge \citep{MICCAI2017} provides three sub-databases which are CVC-ClinicDB, ETIS-Larib and ASU-Mayo Clinic polyp database and which can be used for polyp detection and localization. 
CVC-ClinicDB is a cooperative work of the Hospital Clinic and the Computer Vision Center, Barcelona, Spain. It contains 612 images from 31 different sequences. Each image has its annotated ground truth associated, covering the polyp \citep{CVCClinicDB}. 
ETIS-Larib is a database consisting of 300 frames with polyps extracted from colonoscopy videos. Frames and their ground truths are provided by ETIS laboratory, ENSEA, University of Cergy-Pontoise, France \citep{ETISLaribDB}.
The ASU-Mayo Clinic polyp database was acquired as a cooperative work of Arizona State University and Mayo Clinic, USA. It consists of 20 short colonoscopy videos (22,701 frames) with different resolution ranges and different area coverage values for training purposes. Each frame in its training dataset comes with a ground truth image or a binary mask that indicates the polyp region. In addition, it contains 18 videos without annotation for testing purposes \citep{ASUMayoDataset}.

\item The Hamlyn Centre Laparoscopic/Endoscopic Video Dataset consists of 37 subsets. The Gastrointestinal Endoscopic Dataset includes 10 videos and consists of 7,894 images with a size of 2.5 GB which were collected during standard gastrointestinal examinations. The dataset includes images for polyp detection, localization and optical biopsy retargeting. Apart from endoscopy dataset for depth estimation, one of the laparoscopy datasets contains ~40,000 pairs of rectified stereo images collected in partial nephrectomy using Da Vinci surgery robot. Its primary use has been training and testing deep learning networks for disparity (inverse depth) estimation \citep{Ye2016OnlineTA}, 
\citep{ye2017self}.
\item ROBUST-MIS Challenge provides a dataset which was created in the Heidelberg University Hospital, Germany during rectal resection and proctocolectomy surgeries. Videos from 30 minimal invasive surgical procedures with three different types of surgery and extracted 10,040 standard endoscopic image frames from these 30 procedures performed a basis for this challenge. These images were acquired using a laparoscopic camera (Karl Storz Image 1) with a 30\textdegree\ optic and a resolution of 1920x1080 pixels. The images are, then, downscaled to 960x540 pixels and annotated with numbers showing the absence or presence of medical instruments \citep{RobustMISChallenge}.
\end{itemize}
\newpage
\section{Camera Calibration}
For the coordinate transformation between robot pose data and capsule cameras, hand-eye calibration procedure was repeated with two different checkerboards:  one with \( 2\times \SI{2}{\milli\meter}\) squares and one with \( 1.5\times \SI{1.5}{\milli\meter}\) squares, both patterns with \(8\times 7\) squares in total. Four images of each checkerboard were acquired from different camera poses.  
For the pose conversions, only the checkerboard images from Mirocam capsule was used, with the support structure being the same for both capsules (Pillcam and Mirocam). Similarly, to calculate the transformation between the gripper holding HighCam-LowCam and the camera positions, same procedure was repeated by using the checkerboard squares with \( 10.2\times \SI{10.2}{\milli\meter}\). 

\begin{figure*}[h!]
\centering
{\includegraphics[width=\columnwidth]{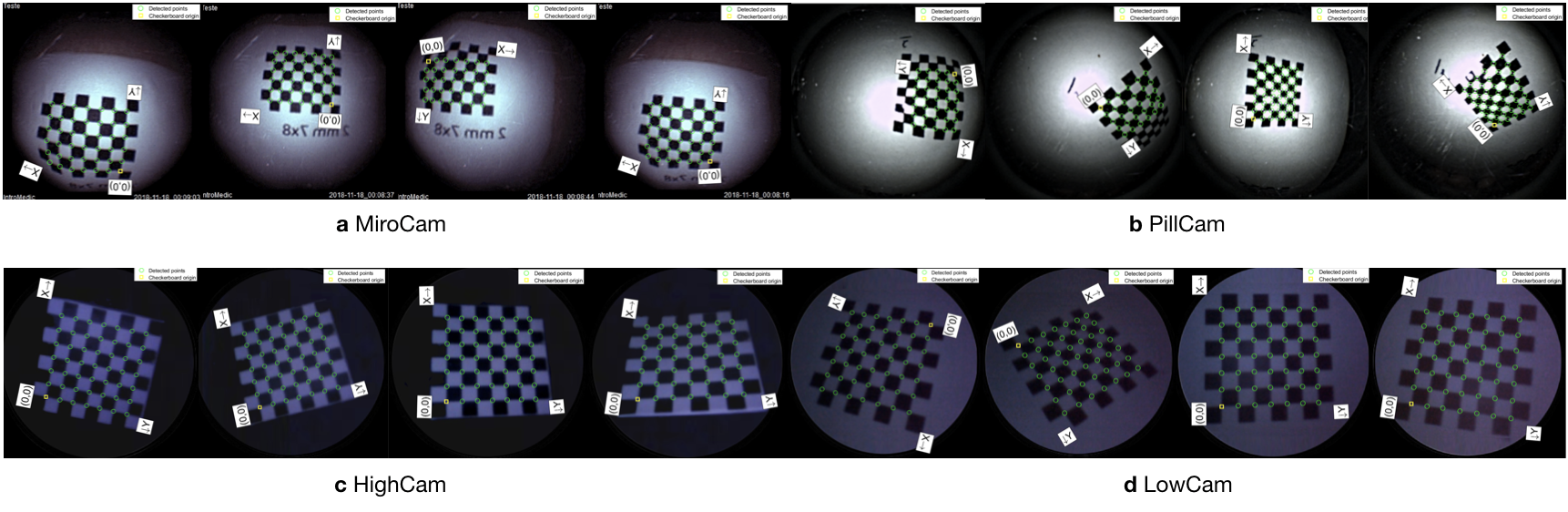}}
\caption{\textbf{Camera Intrinsic-Extrinsic Calibration Images.} Examples of planar checkerboard calibration images obtained by \textbf{a} MiroCam, \textbf{b} PillCam, \textbf{c}  HighCam and \textbf{d} LowCam. The chessboards are printed with a laser printer and then glued on the surface of a planar glass to ensure the planarity of the pattern. Since the dataset is recorded in dark room, chessboard images are taken in same environmental conditions.}
\label{fig:MirocamCalib2mm}
\end{figure*}

The Tsai and Lenz algorithm\citep{HandEyeCalibration1989} was tested with 24 combinations of the 4 chessboard images in Fig. \ref{fig:MirocamCalib2mm}. 
The transformation between a point \(\bf{X}_c\) in the reference frame of the camera  and a point \(\bf{X}_g\) in the reference frame of the gripper is, thus, given by 
\begin{equation}
\bf{X}_g = \bf{R}^c_g \, \bf{X}_c + \bf{t}^c_g
\label{eqn:transformation}
\end{equation} 
with the rotation matrices and translation vectors given in Table ~\ref{tab:hand_eye}.

\begin{table}[!ht]
 \caption{\textbf{Robot Pose to Camera Transformation.} The rotation matrices and translation vectors for MiroCAM, HighCam and LowCam to apply the transformations given in Eqn. \ref{eqn:transformation}. These values are provided as a .txt file and as a .mat file in the calibration folders of the EndoSLAM Dataset.}
     \centering
     \begin{tabular}{ccc}
         Camera & Rotation $\bf{R}^c_g $ & Translation 
         $ \bf{t}^c_g$(mm)
         \\ \midrule\midrule[.1em] 
          MiroCam &
 $\begin{bmatrix}
 -0.9366 & -0.3242 & -0.1325 \\ 0.1738 & -0.1017 & -0.9795 \\ 0.3041 & -0.9405 & 0.1516
 \end{bmatrix}$
 &   $\begin{bmatrix}
 2.9793 \\ -27.0224 \\ 72.1070 
 \end{bmatrix}$ 
 \\ \midrule 
 HighCam &
 $\begin{bmatrix}
 0.9463 & -0.0921 & -0.3098 \\ -0.1389 & 0.7495 & -0.6472 \\ 0.2918 & -0.6555 & 0.8965
 \end{bmatrix}$
 &   $\begin{bmatrix}
 -46.2017 \\ 20.9074 \\ 94.6349
 \end{bmatrix}$ 

 \\\midrule
  LowCam &
 $\begin{bmatrix}
 0.8294 & 0.5577 & 0.0322 \\ -0.5586 & 0.8286 & 0.0379 \\ -0.0056 & -0.0495 & 0.9988
 \end{bmatrix}$
 &   $\begin{bmatrix}
 6.0169 \\ 39.5114 \\ 101.6431 
 \end{bmatrix}$ 
 \\
 \midrule[.1em]
 \end{tabular}
 \label{tab:hand_eye}
 \end{table} 
 For the detailed description of vision based calibration technique and its written codes in MATLAB R2020a, visit { \textit{\url{https://github.com/CapsuleEndoscope/EndoSLAM}}}.
 Calibration was performed for both the Mirocam and Pillcam capsules, using images of a planar checkerboard with \(8\times 7\) squares of dimension \( 2\times \SI{2}{\milli\meter}\) and also for HighCam and LowCam using \(8\times 7\) squares of dimension \( 12.8\times \SI{12.8}{\milli\meter}\). The calibration checkerboard was printed using a laser printer and then glued on the surface of a glass plate to ensure the planarity of the pattern. 

The practical distance and orientation range at which the calibration checkerboard can be placed is limited by the low resolution and depth of field of the cameras. For each camera, 10 calibration images were used with the pattern placed at different poses. The average distance from the camera  was approximately \SI{10}{\milli\meter} for capsule cameras.
Fig. \ref{fig:MirocamCalib2mm} show examples of some of the calibration images. 

\begin{figure*}[!t]
\centering
{\includegraphics[width=\columnwidth]{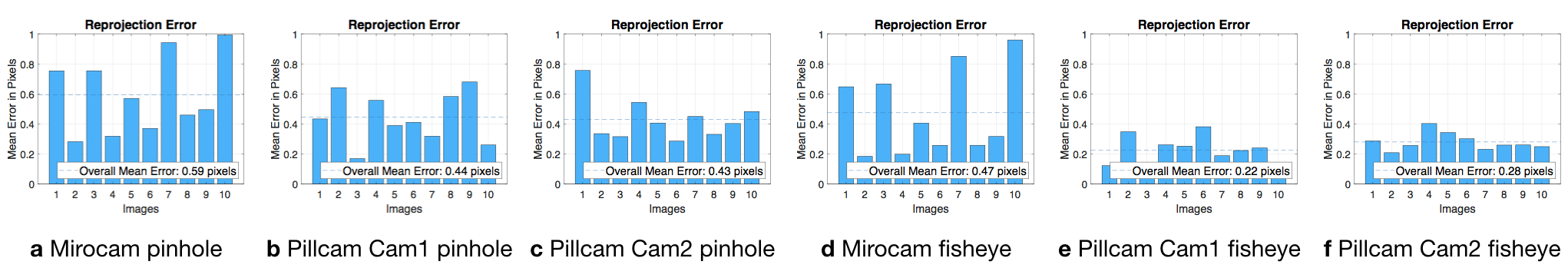}}
\caption{\textbf{Reprojection errors associated with the camera calibrations.} The reprojection errors under pinhole camera assumption for \textbf{a} Mirocam, \textbf{b} Pillcam with a front-facing (Cam1) \textbf{c} Pillcam with a backwards-facing (Cam2) camera. \textbf{d}-\textbf{f} Reprojection errors for the same devices under the fisheye model assumptions.}
\label{fig:CalibReprojErrors}
\end{figure*}

\begin{figure*}[!t]
\centering
{\includegraphics[width=\columnwidth]{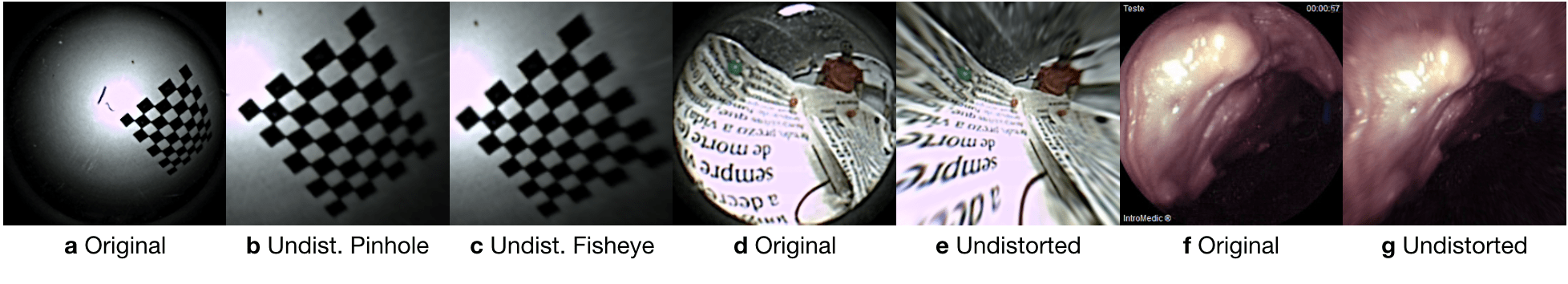}}
\caption{\textbf{Correction of lens distortions.} 
Examples to correct the lens distortions via camera parameters given in Table  \ref{tab:LowHighCamCalib} for the images acquired by PillCam and MiroCam. \textbf{a} Original 8$\times$7 checkerboard image with 2$\times$2mm squares obtained by PillCam, \textbf{b} Undistorted checkerboard image with pinhole calibration parameters, \textbf{c} Undistorted checkerboard image with fisheye parameters, \textbf{d} Newspaper image which is rich in texture details taken by frontal camera of PillCam \textbf{e} Undistorted counterpart of newspaper image with the calculated parameters under fisheye camera assumption. Similarly, \textbf{f} Original Colon-III image of MiroCam and \textbf{g} Undistorted version by the parameters of fisheye calibration model.}
\label{fig:CalibUndistort}
\end{figure*}

Table \ref{tab:LowHighCamCalib} summarizes the estimated intrinsic parameters for each of the calibration models. Note that the Pillcam capsule has both a front-facing and a backwards-facing camera. That dual camera setup can lead to interesting novel visual SLAM approaches that makes simultaneous use of front- and backwards-facing cameras. 
The reprojection errors associated with each calibration can be seen in Fig. \ref{fig:CalibReprojErrors}. Fig. \ref{fig:CalibUndistort} shows some examples for the application of estimated intrinsic parameters to correct the lens distortion effects on capsule images to produce undistorted pinhole images. Note that eight images were used due to the limited operational volume to place the chessboard. Outside that volume the images become either less focused/blurry or if too far the corners are undetectable due to low image quality and resolution.
{\small
\begin{table*}
\caption{\textbf{Intrinsic parameters for MiroCam, PillCam, HighCam and LowCam under the pinhole assumption} Since the effects mentioned in Sec. \ref{ssec:datasetaugmentation} are directly related with camera lens properties, the intrinsic parameters representing new images are also changing compatible with modifications. For the mixture of Gaussian blur, vignetting and resizing effects, the new parameters are given in High-Modified and Low-Modified columns.}
\begin{tabular}{lllrrrrrrr}                              
& & &  &&& &  & \multicolumn{2}{c}{PillCam}  \\ \cmidrule(lr){9-10} 
& & &  {HighCam}  & { LowCam} & { High-Modified} & { Low-Modified}& {MiroCam}& {Cam1} & { Cam2}  \\
  \\ \midrule\midrule[.1em]
& &  H x W  &  \(480 \times 640\) &  \(480 \times 640\) & \(400 \times 400\) & \(250 \times 250\) & \(320 \times 320\) & {\(256 \times 256\)} &{\(256 \times 256\)}\\ \midrule[.1em] 
\parbox[c]{2mm}{\multirow{8}{*}{\rotatebox[origin=c]{90}{Pinhole}}}
 & \multirow{2}{*}{\shortstack{Focal \\ length}} & \(f_x\) &  957.4119
 & 816.8598 & 603.5105 & 317.6319 &  156.0418 & 74.2002 & 76.0535 \\ \cmidrule(lr){3-10} 
 &                               & \(f_y\) & 959.3861 & 814.8223 & 807.6887 & 423.1068 & 155.7529 & 74.4184 & 75.4967  \\ \cmidrule(lr){2-10}
 & Skew                                      & \(s\) &  5.6242 & 0.2072 & 4.2831 & -0.3334 & 0 & 0 & 0  \\ \cmidrule(lr){2-10}
 & \multirow{2}{*}{\shortstack{Optical \\ center}}           & \(c_x\)  & 282.1921 & 308.2864 & 173.7160 & 121.3764 & 178.5604 & 129.9724 & 130.9419\\ \cmidrule(lr){3-10}
 &                                           & \(c_y\)  & 170.7316 & 158.3971 & 133.7022 & 82.5754  & 181.8043 & 129.1209 & 128.4882 \\ \cmidrule(lr){2-10}
 & \multirow{3}{*}{\shortstack{Radial \\ dist. coef.}} & \(k_1\)  & 0.2533 & 0.2345 & 0.2645 & 0.2265  & -0.2486 & 0.1994 & 0.1985 \\ \cmidrule(lr){3-10}
 &                                           & \(k_2\)  & -0.2085 & -0.7908 & -0.4186 & -0.8877    & 0.0614  & -0.1279 & -0.1317  \\ \midrule[.1em]               
 \bottomrule                
\end{tabular}

\label{tab:LowHighCamCalib}
\end{table*}}

\clearpage
\section{Temporal-Synchronization}
\begin{figure*}[!th]
\centering
{\includegraphics[width=\columnwidth]{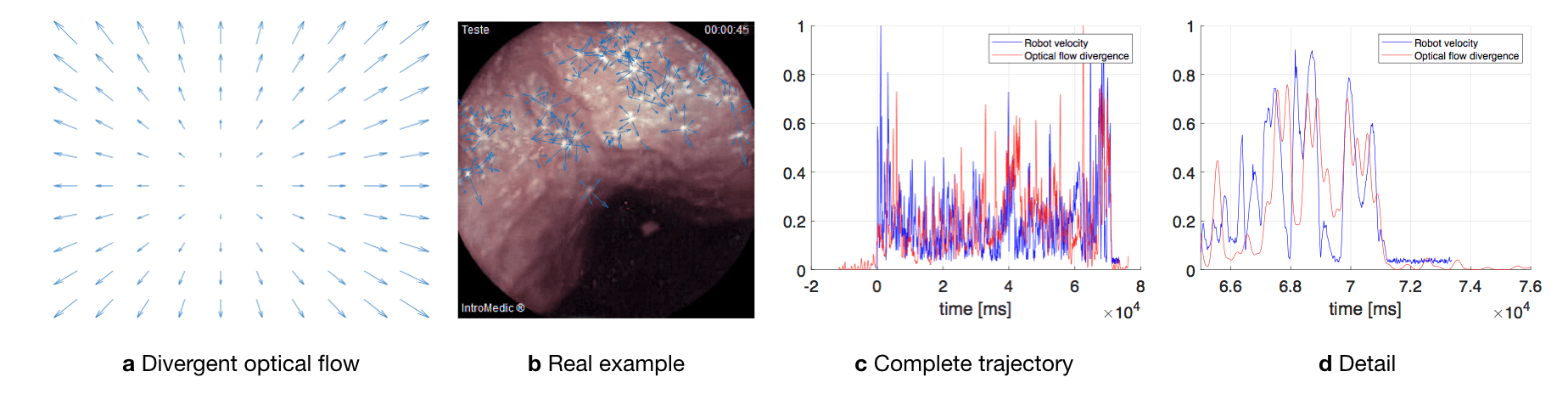}}
\caption{\textbf{Optical flow and temporal synchronization of robot data and images.} The divergence of the optical flow field is used to estimate the forward motion of the camera. \textbf{a} The divergent optical flow field obtained as a pinhole camera moves towards a frontal-parallel plane.  \textbf{b} Image from MiroCam, of divergent optical flow detected in image areas of high contrast.  \textbf{c} The camera velocity for the sixth sequence of MiroCam record, obtained from the robot position data (low pass filter was used to reduce noise), and estimated from the divergence of the optical flow field in the images. The vertical scale is normalized for both measurements. The signals are correlated and aligned to obtain temporal synchronization. \textbf{d} Plot corresponding to the end of the trajectory, when the camera movement stops. }
\label{fig:MirocamTempSyncFrame}
\end{figure*}

The apparent velocity field in the image, i.e. the optical flow, is the projection of the 3D velocity of the scene (w.r.t. the camera), and is, in general,  dependent  on scene depth~\citep{honegger2012real}.  In the  dataset,  the camera moves, roughly, in a straight line along its optical axis and with limited rotation, for the most part of the trajectory. The scene is also relatively uniform and symmetric, in terms of relative depth. To estimate this "forward" motion, the divergence of the flow vector  field can be used~\citep{ho2017distance,mccarthy2008robust}. To provide insight, Fig. \ref{fig:MirocamTempSyncFrame}a shows the divergent optical flow field that would be obtain as a pinhole camera moves towards a frontal-parallel plane.  Fig. \ref{fig:MirocamTempSyncFrame}b shows an example, obtained from the dataset, of divergent optical flow in image areas of high contrast.
 
 The divergence at an image point \( (x,y) \) is given by
 \begin{equation}
     D(x,y)= \frac{\partial u(x,y)}{\partial x} + \frac{\partial v(x,y)}{\partial v}
 \end{equation}
 where \(u\) and \(v\) are the velocity components of the optical flow field and \(\partial\) denotes a partial derivative. The divergence measurement is averaged across all points, yielding a single estimated value for each image. MATLAB was used to compute the optical flow and the flow divergence, using the method in~\citep{lucas1981iterative}.

The robot encoder data provides the camera pose along the trajectory. The linear velocity is computed by applying a finite difference on the position data, followed by a low pass filter~\citep{liu2002velocity,puglisi2015velocity}. Specifically, the camera velocity   \(v\) is obtained from 
\begin{equation}
    v_k=|| \frac{\bf{X}_k - \bf{X}_{k-1}}{T} ||
\end{equation}
where  \(\bf{X}\) denotes the 3D position  vector, \(T\) denotes the sampling period, and the subscript  \(k\) indexes the sample instant. A low pass Butterworth filter (with a cutoff frequency of 300Hz) is  then applied to the velocity measurement in order to reduce noise.
 
The optical flow divergence and camera velocity measurements are correlated along the time axis to determine the best alignment. As an example, Fig. \ref{fig:MirocamTempSyncFrame}c shows the camera velocity (magnitude) during sequence/experiment for the sixth sequence of MiroCam record, 
calculated from the robot position data and estimated from the divergence of the optical flow field in the images. Both signals are shown already synchronized. Fig. \ref{fig:MirocamTempSyncFrame} d shows a detail of the plot corresponding to the end of the trajectory, when the camera stops moving.

 Table \ref{tab:TempSyncHighLow} and \ref{tab:TempSync} summarize the temporal synchronization for all trajectories in the dataset. They provide a correspondence between the start frame of each sequence and the matching sampling instant of the robot pose data. 
\begin{table}[!th]
\centering
\caption{\textbf{Temporal synchronization.} Correspondence, for each sequence of each organ, between the first frame of the trajectory for both HighCam and LowCam and the matching sample instant of the robot data with 1kHz recording frequency.}
\resizebox{0.5\columnwidth}{!}{
\begin{tabular}{ccccccc}
 & & \multicolumn{2}{c}{Camera} & \multicolumn{2}{c}{Robot} \\ \cmidrule(lr){3-4} \cmidrule(lr){5-6}
{Organ} & {Trajectory} & \vtop{\hbox{\strut HighCam}\hbox{\strut Start Frame}} & \vtop{\hbox{\strut LowCam}\hbox{\strut Start Frame}} & \vtop{\hbox{\strut HighCam}\hbox{\strut Sample}} &  \vtop{\hbox{\strut LowCam}\hbox{\strut Sample}}  \\  \midrule\midrule[.1em]
\parbox[t]{2mm}{\multirow{7}{*}{\rotatebox[origin=c]{90}{Colon-IV}}}  
   & 1 & 741 & 393  & 35,295 & 15,845 \\ \cmidrule(lr){2-6}
   & 2 & 44 & 128  & 2,561 & 2,561 \\ \cmidrule(lr){2-6}
   & 3 & 69 & 82  & 3,975 & 3,975 \\ \cmidrule(lr){2-6}
   & 4 & 138 & 120  & 15,792 & 15,092 \\ \cmidrule(lr){2-6}
   & 5 & 99 & 144  & 1,270 &  3,270\\  \midrule[.1em]  
\parbox[t]{2mm}{\multirow{5}{*}{\rotatebox[origin=c]{90}{SmallIntestine}}}
   & 1 & 149 & 95  & 5,162 & 4,512 \\ \cmidrule(lr){2-6}
   & 2 & 133 & 112  & 4,913 & 2,763 \\ \cmidrule(lr){2-6}
   & 3 & 186 & 144  & 6,095 &7,845  \\ \cmidrule(lr){2-6}
   & 4 & 121 & 79  & 3,205 & 3,205 \\ \cmidrule(lr){2-6}
   & 5 & 138 & 105  & 3,807 & 3,307 \\  \midrule[.1em]  
\parbox[t]{2mm}{\multirow{5}{*}{\rotatebox[origin=c]{90}{Stom-I}}} 
   & 1 & 60 & 135  & 4,443 &  8,093\\ \cmidrule(lr){2-6}
   & 2 & 111 & 144  & 4,177 &  2,277\\ \cmidrule(lr){2-6}
   & 3 & 71 & 447  & 6,058 & 19,008 \\ \cmidrule(lr){2-6}
   & 4 & 47 & 316  & 2,839 & 13,289 \\ \midrule[.1em]
\parbox[t]{2mm}{\multirow{5}{*}{\rotatebox[origin=c]{90}{Stom-II}}}
   & 1 & 255 & 125  & 9,641 & 5,141 \\ \cmidrule(lr){2-6}
   & 2 & 1 & 2  & 3,358 &  3,358\\ \cmidrule(lr){2-6}
   & 3 & 150 & 83  & 5,797 &  2,247\\ \cmidrule(lr){2-6}
   & 4 & 78 & 85  & 2,742 &  4,192\\ \midrule[.1em]
\parbox[t]{2mm}{\multirow{5}{*}{\rotatebox[origin=c]{90}{Stom-III}}}
   & 1 & 195 & 89  & 6,746 &  2,846\\ \cmidrule(lr){2-6}
   & 2 & 302 & 108  & 1,523 & 2,725 \\ \cmidrule(lr){2-6}
   & 3 & 387 & 105  & 17,261 & 2,861 \\ \cmidrule(lr){2-6}
   & 4 & 125 & 60  & 4,451 & 2,101 \\ \midrule[.1em]
\end{tabular}
}

\label{tab:TempSyncHighLow}
\end{table}

\begin{table}[h]
\centering
\caption{\textbf{Temporal synchronization.} Correspondence, for each sequence, between the first frame of the trajectory and the matching sample instant (sample number) of the robot data. Note that, in the Pillcam capsule, Cam1 (front facing camera) and  Cam2 (backward facing camera) trigger alternatively, one after the other, with equally spaced time intervals. The values indicated in the table correspond to Cam1.}
\scriptsize
\begin{tabular}{cccccc}
 & & \multicolumn{2}{c}{Camera} & \multicolumn{2}{c}{Robot} \\ \cmidrule(lr){3-4} \cmidrule(lr){5-6} 
\multicolumn{2}{c}{Sequence} & start frame     & framerate    & sample instant   & sampl. freq.  \\  \midrule\midrule[.1em]
\multirow{6}{*}{Mirocam} 
   & 1 & 336 & 3 fps  & 72,050 & 1kHz \\ \cmidrule(lr){2-6}
   & 2 & 153 & 3 fps  & 961 & 1kHz \\ \cmidrule(lr){2-6}
   & 3 & 321 & 3 fps  & 47,667 & 1kHz \\ \cmidrule(lr){2-6}
   & 4 & 143 & 3 fps  & 33,943 & 1kHz \\ \cmidrule(lr){2-6}
   & 5 & 254 & 3 fps  & 2,886 & 1kHz \\ \cmidrule(lr){2-6}
   & 6 & 134 & 3 fps  & 3,044 & 1kHz \\  \midrule[.1em]  
\multirow{2}{*}{Pillcam} 
   & "L" & 1,127 & 0.117 fps  & 15,800 & 1kHz \\ \cmidrule(lr){2-6}
   & "Z" & 815 & 0.117 fps  & 11,650 & 1kHz \\  \midrule[.1em]   
\end{tabular}

\label{tab:TempSync}
\end{table}

\clearpage
\section{Data Tree Structure}
\begin{figure}[!ht] 
 \centering
 \includegraphics[width = 0.6\columnwidth]{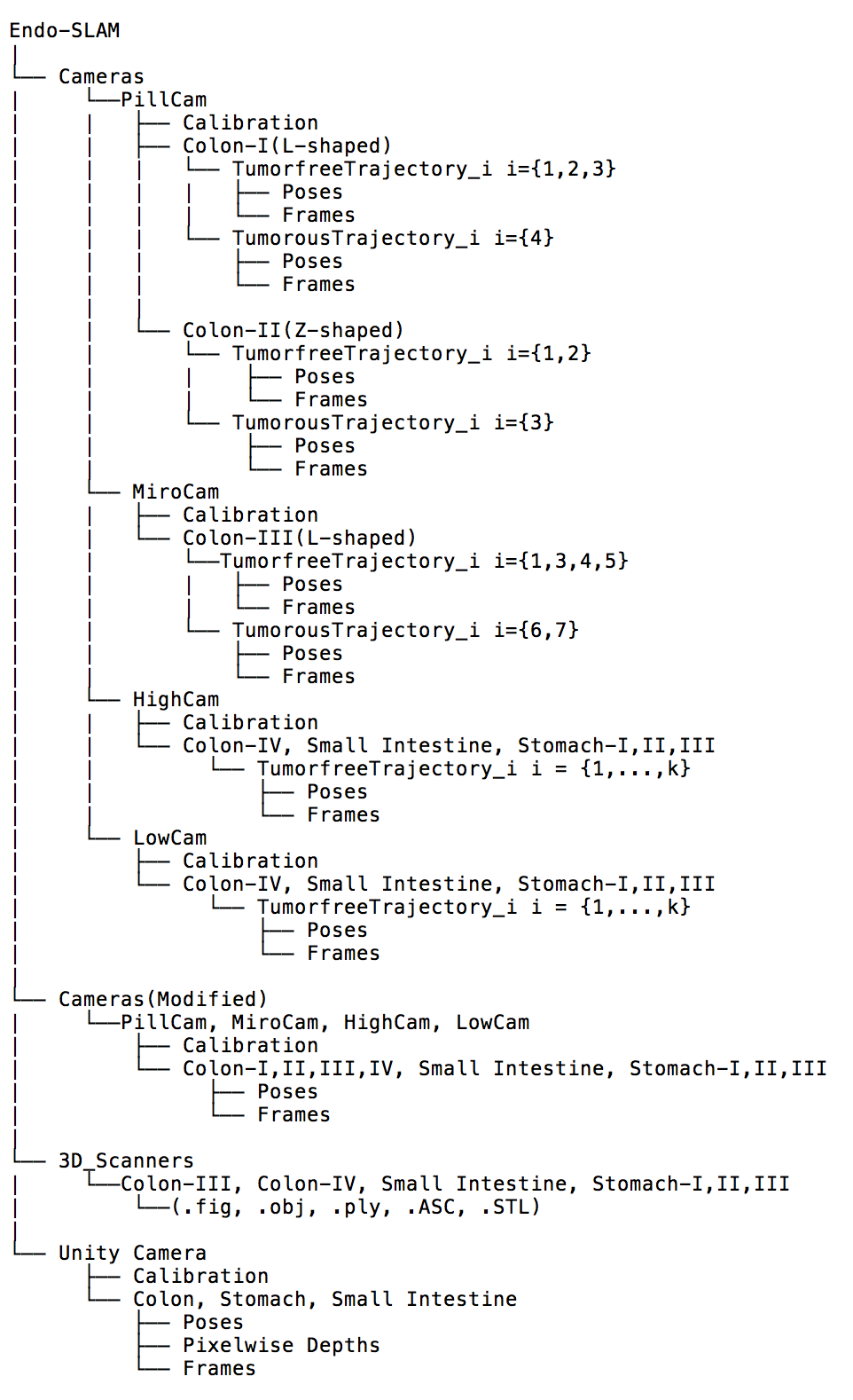}
  \caption{
  \textbf{Data Tree.} 
  EndoSLAM dataset is divided into four main parts: Cameras, Cameras (Modified), 3D\textunderscore Scanners and Unity Camera.
  Cameras and Cameras (Modified) folders include four subfolders as PillCam, MiroCam,  HighCam and LowCam. Each of them branches out into calibration and organs subfolders. Calibration subfolder comprises intrinsic-extrinsic camera parameters and corresponding calibration sessions whereas organs subfolder includes images and poses of each trajectories. Apart from Cameras section, modified part includes sample sub-trajectories exemplifying the effects of image modification functions such as fish-eye, Gaussian blur, vignetting, resizing, depth of fields. 3D\textunderscore Scanners folder consists of 
  reconstructed 3D figures (.fig), point cloud data (.ply), surface geometry of three-dimensional objects without any color or texture representations (.STL), the position of each vertex representing 3D geometry(.obj) and ASCII formatted point cloud data(.ASC). Finally Unity Camera folder includes synthetically generated images, pixelwise depths and corresponding poses.}
  \label{fig:folder}
\end{figure}

\clearpage
\section{Dataset Analysis}
\begin{figure*}[!ht]
    \centering
    \includegraphics[width = \columnwidth]{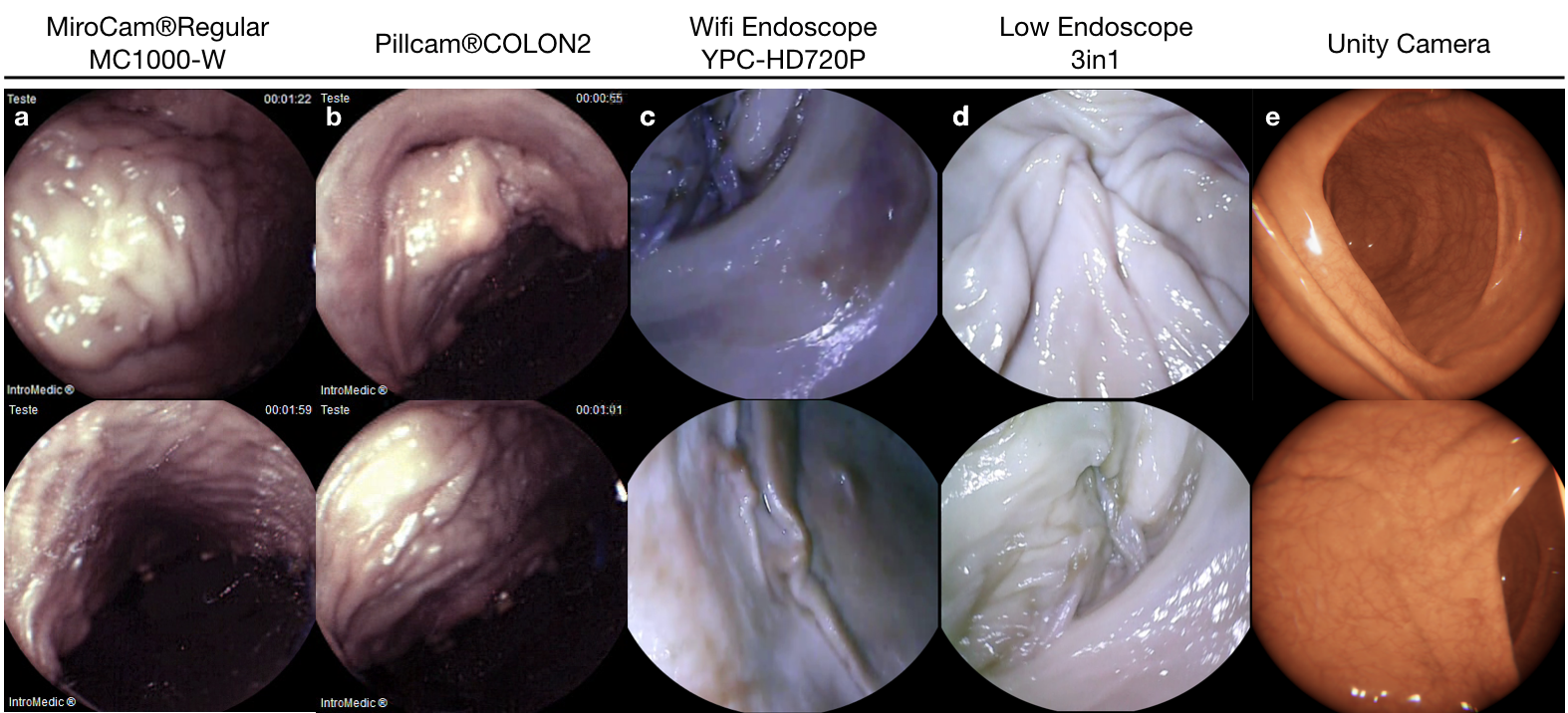}
    \caption{\textbf{Sample frames from EndoSLAM Dataset.} Images are acquired by \textbf{a} MiroCam capsule endoscope, \textbf{b} Frontal camera of a PillCam, \textbf{c} HighCam, \textbf{d} LowCam, and \textbf{e} virtually generated UnityCam. The ex-vivo part of the dataset offers opportunity to test the robustness of pose estimation algorithms with images coming from various endoscope camera. Since EndoSLAM dataset contains real and simulated frames, it is also a suitable platform to develop domain adaptation algorithms.  
    }
    \label{fig:samptraj2}
\end{figure*} 
{\centering
\begin{table*}[!hbt]
\caption{\textbf{Motion Analysis.} Statistics for robot poses matching with frames of HighCam. For all trajectories of each organ, counts of robot sample instances, mean, first quantile(1st QT), median, third quantile(3rd QT), minimum, maximum speed[mm/s] values are given.} 
{\small\centering
\begin{tabular}{ccccccc}
\parbox[t]{2mm}{\multirow{20}{*}{\rotatebox[origin=c]{90}{Speed}}}

     &&{Stomach-I} & {Stomach-II} & {Stomach-III} & {Small Intestine} & {Colon-IV} \\
     \midrule\midrule[.1em]
     &{frame count} & 4695  & 3302 & 3230 & 6487 & 3697\\
     \cmidrule(lr){2-7}
     &{mean[mm/s]}&18.256&19.471 &20.031&16.764 &17.123\\ 
      \cmidrule(lr){2-7}
      &{std[mm/s]}&22.497&16.809 &16.697&14.210 &12.660\\ 
          \cmidrule(lr){2-7}
&{1st QT}&5.931&7.606 &7.055&5.684 &7.658\\ 
    \cmidrule(lr){2-7}
&{median}&14.642& 16.021&16.489&13.849 &15.096\\ 
    \cmidrule(lr){2-7}
&{3rd QT}&25.32&26.64 &28.68&24.342 &24.324\\
    \cmidrule(lr){2-7}
&{min[mm/s]} &0.02&0.028 &0.02& 0&0.007\\
    \cmidrule(lr){2-7}
&{max[mm/s]}&25.32&140.898 &116.984&104.08 &104.759\\
\midrule[.1em] 
\parbox[t]{2mm}{\multirow{10}{*}{\rotatebox[origin=c]{90}{Acceleration}}}
     &{mean[mm/s]}&359.843&382.928 &383.829& 328.568&326.241\\ 
      \cmidrule(lr){2-7}
      &{std}&450.939&337.08 &336.098&284.423 &257.08\\ 
          \cmidrule(lr){2-7}
&{1st QT}&110.408& 140.982&111.71& 103.129&122.807\\ 
    \cmidrule(lr){2-7}
&{median}&284.729&314.784 &315.012& 269.316&283.728\\ 
    \cmidrule(lr){2-7}
&{3rd QT}&501.31&528.799 &556.451&477.991 &469.113\\
    \cmidrule(lr){2-7}
    &{min[mm/s]}&0.4& 0.0&0.0&0.0 &0.015\\
\cmidrule(lr){2-7}
&{max[mm/s]}&14,680.15& 2,817.962&2,339.683&2,079.994 &2,095.182\\
\midrule[.1em] 
\end{tabular}

\label{tab:trajSpeedAccHigh}
}

\end{table*}
}
{\centering
\begin{table*}[!hbt]
\caption{\textbf{Motion Analysis.}  Statistics for robot poses matching with frames of LowCam. For all trajectories of each organ, counts of robot sample instances, mean, first quantile(1st QT), median, third quantile(3rd QT), minimum, maximum speed[mm/s] values are given.} 
{\small\centering
\begin{tabular}{ccccccc}
\parbox[t]{2mm}{\multirow{20}{*}{\rotatebox[origin=c]{90}{Speed}}}

     &&{Stomach-I} & {Stomach-II} & {Stomach-III} & {Small Intestine} & {Colon-IV} \\
     \midrule\midrule[.1em]
     &{frame count} & 2302  & 2799 & 3900 & 5098 & 3857\\
     \cmidrule(lr){2-7}
     &{mean[mm/s]}&15.599&18.928 &25.97&17.918 &17.144\\ 
      \cmidrule(lr){2-7}
      &{std[mm/s]}&12.855&14.431 &21.564&14.764 &12.882\\ 
          \cmidrule(lr){2-7}
&{1st QT}&5.407&8.259 &10.789&6.126 &7.401\\ 
    \cmidrule(lr){2-7}
&{median}&13.18& 15.871 &21.284&15.322 &15.148\\ 
    \cmidrule(lr){2-7}
&{3rd QT}&22.956.&26.436 &35.763&26.146 &24.455\\
    \cmidrule(lr){2-7}
&{min[mm/s]} &0.02&0.0 &0.02& 0.0&0.028\\
    \cmidrule(lr){2-7}
&{max[mm/s]}&79.042&103.254 &286.68&97.315 &106.271\\
\midrule[.1em] 
\parbox[t]{2mm}{\multirow{10}{*}{\rotatebox[origin=c]{90}{Acceleration}}}
     &{mean[mm/s]}&279.254&378.346 &519.361& 334.373&355.941\\ 
      \cmidrule(lr){2-7}
      &{std}&253.777&288.769 &431.327& 295.646 &259.482\\ 
          \cmidrule(lr){2-7}
&{1st QT}&66.573& 164.972&215.786& 119.299&130.256\\ 
    \cmidrule(lr){2-7}
&{median}&221.297&317.344 &425.678& 303.582&291.994\\ 
    \cmidrule(lr){2-7}
&{3rd QT}&428.253&528.695 &715.263&520.839 &482.864\\
    \cmidrule(lr){2-7}
    &{min[mm/s]}&0.4& 0.0&0.0&0.0 &0.015\\
\cmidrule(lr){2-7}
&{max[mm/s]}&1,580.846&2.065,071 &5,733.593&1,946.305 &2,125.42\\
\midrule[.1em] 
\end{tabular}

\label{tab:trajSpeedAccLow}
}

\end{table*}
}

\begin{figure*}[!h]
    \centering
    \includegraphics[width=\columnwidth]{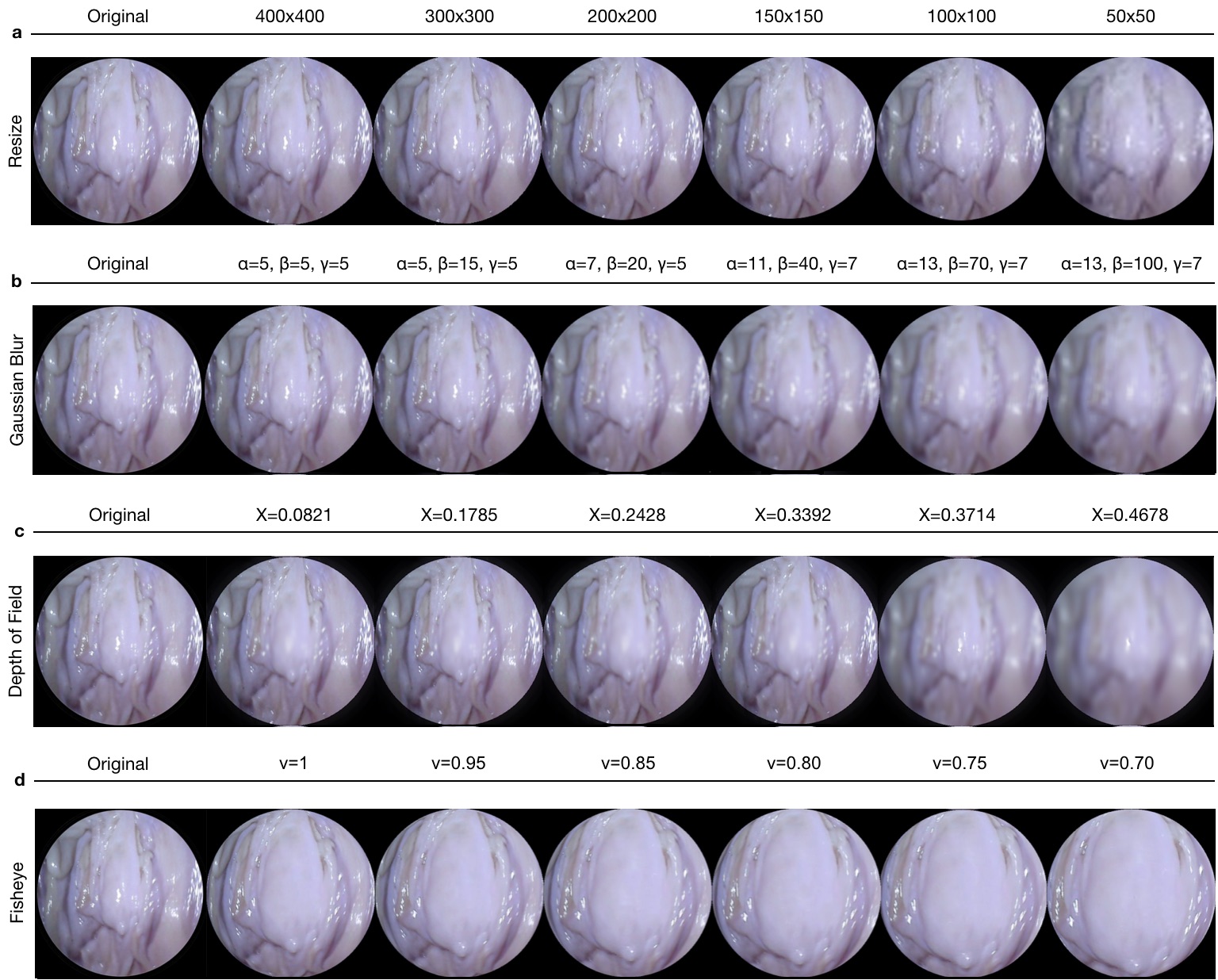}
    \caption{\textbf{Image Modifications.}  \textbf{a Resize} The size of the images, width$\times$height, from left to right is given as 400$\times$400, 300$\times$300, 200$\times$200, 150$\times$150, 100$\times$100 and  50$\times$50, \textbf{b Gaussian Blur} with convolution filter size($\alpha$)  are 5$\times$5,5$\times$5,7$\times$7,11$\times$11,13$\times$13 and 13$\times$13 and standard deviation of Gaussian distribution($\beta$) 5,15,20,40,70,100 and the number o filtering times($\gamma$) 5,5,5,7,7,7.  \textbf{c Depth of Field} effects for the focus positions 0.0821, 0.1785, 0.2428, 0.3392, 0.3714, 0.4678, \textbf{d Fish Eye}  distortion for discarding ratios $\nu$ for 1, 0.95, 0.85, 0.8, 0.75, 0.7.}
    \label{fig:lowCam_effects}
\end{figure*} 
\begin{table}[!h]

\centering
\caption{\textbf{The Classification of Trajectories} The recorded trajectories for each organ divided into two groups based on the tumorous properties of tissue as tumor-containing and tumor-free. Approximately 10\% of all trajectories is tumorous which might be practical for segmentation and disease classification tasks.}
\begin{tabular}{ccc}
Organs & Tumor-free Trajectory \# & Tumor-containing Trajectory \# \\
\midrule\midrule[.1em]
Colon-I & I,II,III, & IV \\ \midrule[.1em]
Colon-II & I,III,IV,V & VI,VII \\ \midrule[.1em]
Colon-III & I,II & III \\ \midrule[.1em]
Colon-IV & I,II,III,IV,V & - \\ \midrule[.1em]
Stomach-I & I,II,III,IV  & - \\ \midrule[.1em]
Stomach-II & I,II,III,IV & - \\ \midrule[.1em]
Stomach-III & I,II,III,IV & - \\ \midrule[.1em]
Small Intestine & I,II,III,IV,V & - \\ \midrule[.1em]
\end{tabular}
\label{table:coltraj}
\end{table}

\begin{table}[ht!]
\normalsize
\caption{\textbf{3D Point Count Data.} The point cloud counts in 3D\textunderscore Scanner folder containing six polygon (.ply) files, for which Colon-III is scanned by Artec 3D Eva with precision \SI{0.1}{\milli\meter}. Colon-IV, Small Intestine and Stomach-I,-II,-III are scanned by Shining 3D EinScan Pro 2x with the precision \SI{0.05}{\milli\meter}.}
\begin{center}
\begin{tabular}{p{3cm} p{3.5cm} p{4cm} p{3cm}} 
Organ & 3D Point Count & Scanner & Precision \\
\hline\hline
Colon-IV & 2,106,046 & 3D EinScan Pro 2x & \SI{0.05}{\milli\meter} \\ \midrule
Small Intestine & 2,193,364 & 3D EinScan Pro 2x & \SI{0.05}{\milli\meter} \\ \midrule
Stomach-I & 2,597,906 & 3D EinScan Pro 2x & \SI{0.05}{\milli\meter} \\ \midrule
Stomach-II & 5,729,625 & 3D EinScan Pro 2x & \SI{0.05}{\milli\meter} \\ \midrule
Stomach-III & 2,234,849 & 3D EinScan Pro 2x & \SI{0.05}{\milli\meter} \\ \midrule
Colon-III & 151,846 & Artec 3D Eva & \SI{0.10}{\milli\meter} \\ \midrule
\end{tabular}
 \label{tab:3Dpoint}
\end{center}
\end{table}

\begin{figure*}[h]
\centering
{\includegraphics[width=\columnwidth]{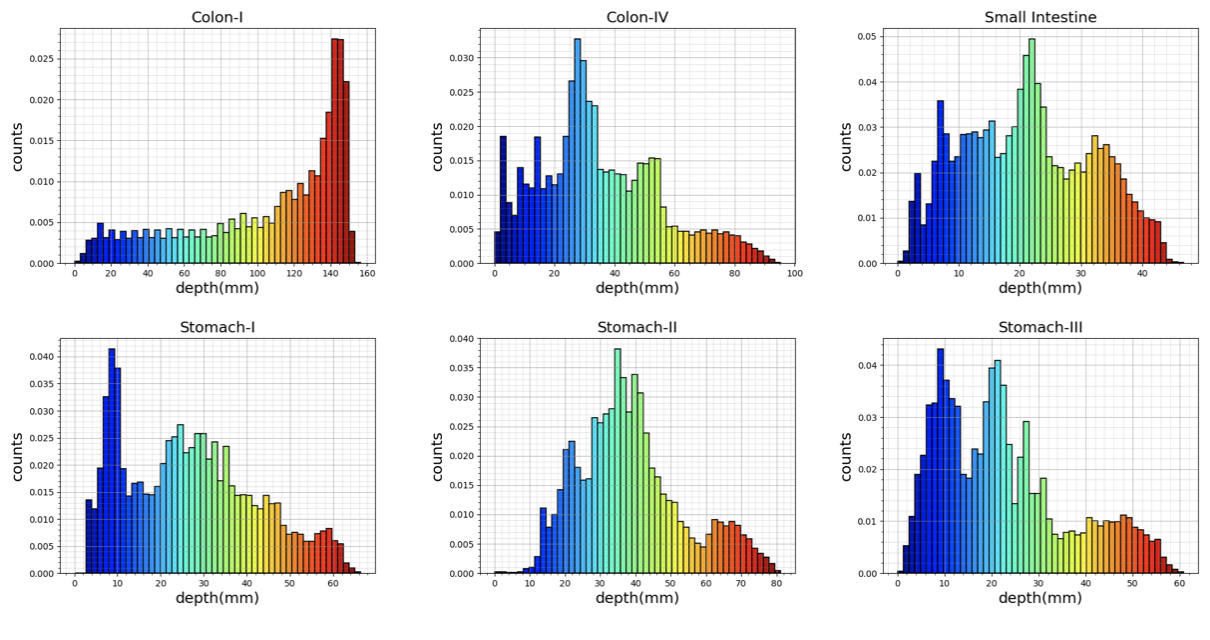}}
\caption{\textbf{Depth Evaluation of Point Cloud Data.} 
The frequency distribution of depth values in mm for \textbf{a} Colon-I scanned by Artec Eva: 3D scanner, \textbf{b} Colon-IV, \textbf{c} Small Intestine, \textbf{d},\textbf{e},\textbf{f} Stomach-I,II,III all scanned by EinScan Pro 2X.}
\label{fig:depthhisto}
\end{figure*}

\begin{figure*}[!t]
\centering
{\includegraphics[width=\columnwidth]{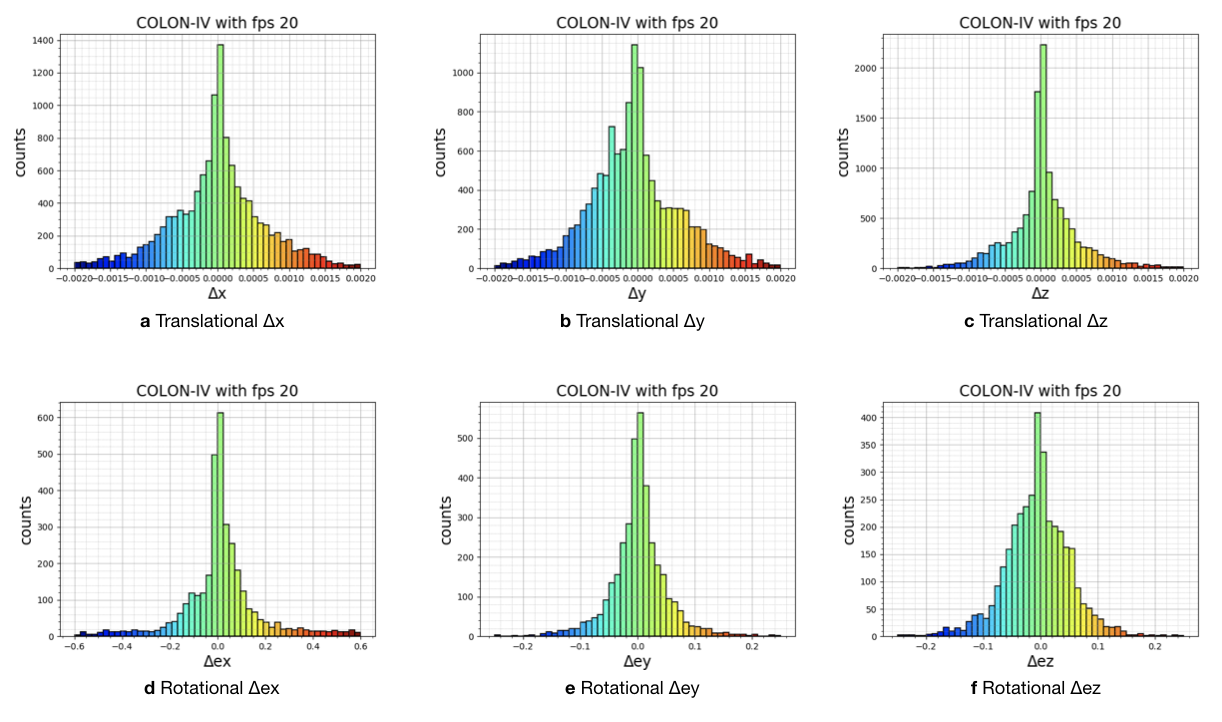}}
\caption{\textbf{Motion Analysis Histograms} The frequency distribution of positional differences between two consecutive frames along the \textbf{a} x, \textbf{b} y, \textbf{c} z axis and the rotational differences in \textbf{d} x, \textbf{e} y,  \textbf{f} z axis in terms of Euler angles are given.}
\label{fig:histo}
\end{figure*}

\clearpage
\section{Results}

\begin{figure*}[!ht]
    \centering
    \includegraphics[width =  \textwidth]{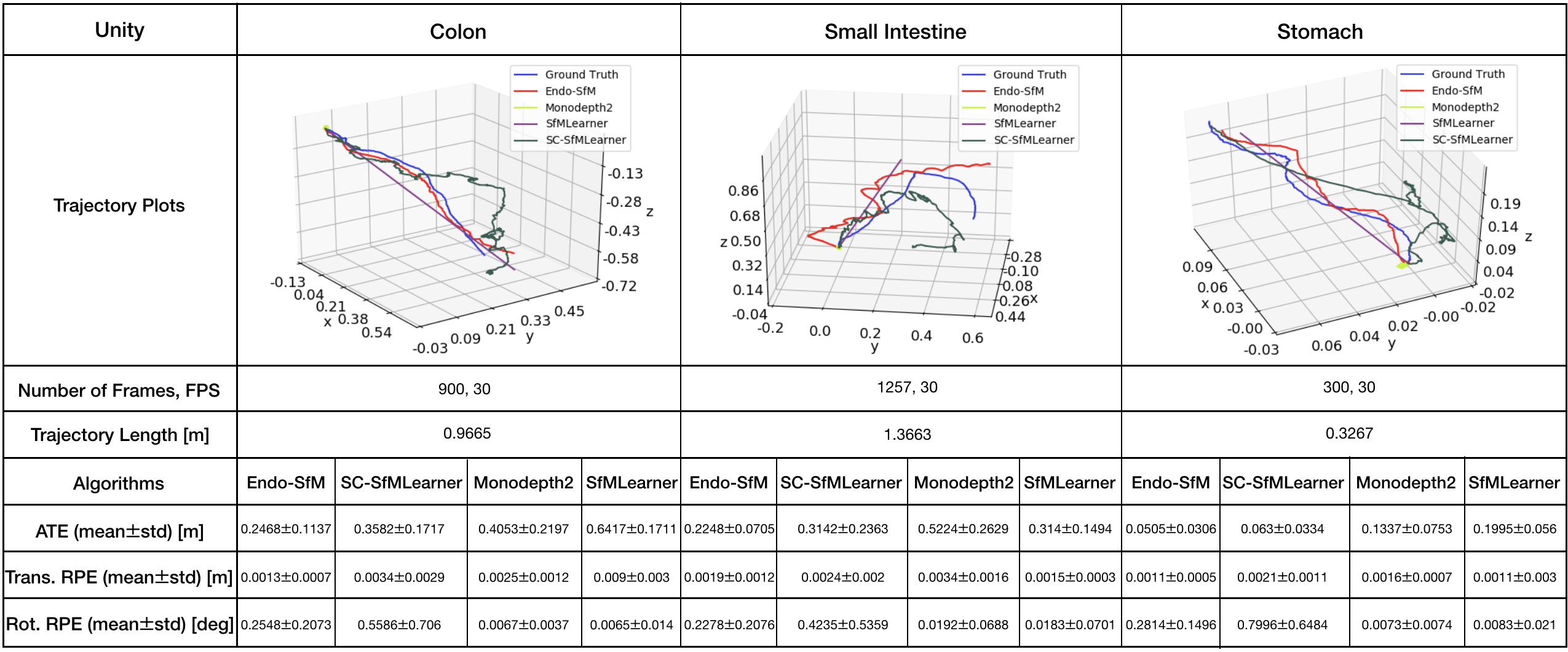}
    \caption{ \textbf{Pose Estimations on Unity Trajectories.} On the contrary to real ex-vivo records, synthetically generated trajectories are more straightforward and easier to follow. This fact results in increase in the performance of all methods. However, SC-SfMLeaner and Endo-SfMLeaner track the route with higher accuracy thanks to the geometry consistency loss. Even if all algorithms trained by the synthetic colon images, Monodepth2 and SfMLearner face with the same problem as in real trajectories. For all synthetically generated trajectories, EndoSfMLearner exhibits lowest mean absolute trajectory error(ATE). Although quantitatively Monodepth2 and SfMLearner have lower rotational error, it cannot be taken into account as performance superiority. Since the rotations cannot be changed frequently and easily while recording clear images in Unity environment, they remain close to identity matrix which is generally predicted by Monodepth2 and SfMLearner.  
    }
    \label{fig:unityposeestimation}
\end{figure*}

\end{document}